\documentclass[11pt]{article}

\usepackage[final]{acl}

\usepackage{times}
\usepackage{latexsym}

\usepackage[T1]{fontenc}
\usepackage[utf8]{inputenc}

\usepackage{microtype}

\usepackage{inconsolata}

\usepackage{graphicx}

\usepackage{capt-of}
\usepackage[utf8]{inputenc} % allow utf-8 input
\usepackage[T1]{fontenc}    % use 8-bit T1 fonts
\usepackage{hyperref}       % hyperlinks
\usepackage{url}            % simple URL typesetting
\usepackage{booktabs}       % professional-quality tables
\usepackage{amsfonts}       % blackboard math symbols
\usepackage{nicefrac}       % compact symbols for 1/2, etc.
\usepackage{microtype}      % microtypography
\usepackage{xcolor}         % colors
\usepackage{subcaption}
\usepackage{pifont} 

\usepackage{enumitem}
\usepackage{graphicx}
\usepackage{amssymb}
\usepackage{amsmath} 
\usepackage{multirow}
\usepackage[table]{xcolor}

\usepackage{tabularx}
\usepackage{cuted}
\newcolumntype{C}{>{\centering\arraybackslash}X}
\usepackage{makecell}

\usepackage{array}
\usepackage{longtable}
\usepackage{algorithm}
\usepackage{algpseudocode}
\usepackage{float}
\usepackage{marvosym}
\usepackage{wrapfig}
\usepackage{fontawesome5}

\usepackage[most]{tcolorbox}
\definecolor{ggreen}{rgb}{0.0, 0.6, 0.0}
\definecolor{rred}{rgb}{0.75, 0.0, 0.0}
\definecolor{bblue}{rgb}{0.13, 0.67, 0.8}

\definecolor{BoxBackground}{RGB}{240, 240, 240} 

\definecolor{BoxFrame}{RGB}{0, 0, 0} 
\definecolor{TitleBackground}{RGB}{0, 0, 0} 
\definecolor{TitleText}{RGB}{255, 255, 255} 
\definecolor{deepgreen}{RGB}{0,100,0}
\definecolor{deepred}{RGB}{199, 0, 0}
\tcbset{
  academicbox/.style={
    boxsep=5pt,
    left=2pt,
    right=2pt,
    bottom=0.5pt,
    boxrule=0.5pt,
    colback=BoxBackground,
    colframe=BoxFrame,
    colbacktitle=TitleBackground,
    coltitle=TitleText,
    enhanced,
    attach boxed title to top left={yshift=-0.1in,xshift=0.1in},
    boxed title style={boxrule=0pt,colframe=white},
    title={#1},
  }
}

\newtcolorbox{AcademicBox}[1][]{academicbox=#1}
\definecolor{SoftBlue}{RGB}{135, 206, 250}  
\definecolor{SoftOrange}{RGB}{255, 224, 178} 
\definecolor{SoftGreen}{RGB}{144, 238, 144}  
\definecolor{CorrectGreen}{RGB}{76, 175, 80} 
\definecolor{ErrorRed}{RGB}{211, 47, 47} 

\usepackage{graphicx}
\usepackage{pifont}

\title{MemReread: Enhancing Agentic Long-Context Reasoning via Memory-Guided Rereading
}

\author{
  \normalfont\normalsize
  \begin{tabular}{@{}c@{}}
    \textbf{%
      Baibei Ji\textsuperscript{1},\quad
      Xiaoyang Weng\textsuperscript{1},\quad
      Juntao Li\textsuperscript{1,}\thanks{Corresponding author.},\quad
      Zecheng Tang\textsuperscript{1},\quad
      Yihang Lou\textsuperscript{2},\quad
      Min Zhang\textsuperscript{1}%
    } \\[2pt]
    \textsuperscript{1}Soochow University \quad
    \textsuperscript{2}Peking University \\[2pt]
    \texttt{\{bbji, xyweng35\}@stu.suda.edu.cn}\quad
    \texttt{\{ljt, minzhang\}@suda.edu.cn} \\ [2pt]
    \faGithub\;\texttt{Code:}\quad
    \href{https://github.com/iiGray/MemReread}{%
      \textcolor{magenta}{%
        \texttt{https://github.com/iiGray/MemReread}%
      }%
    }
  \end{tabular}
}

\begin{document}
\maketitle
\begin{abstract}
% To tackle long-context reasoning tasks without the quadratic complexity of standard attention mechanisms, approaches based on agent memory have emerged, which typically maintain a dynamically updated memory when linearly processing document chunks. To mitigate the potential loss of latent evidence in this memorize-while-reading paradigm, recent works have integrated retrieval modules that allow agents to recall information previously discarded during memory overwriting. However, retrieval-based recall suffers from both evidence loss during memory formation and interference induced by invalid queries. To overcome these limitations, we propose MemReread. Built upon streaming reading, MemReread circumvents intermediate retrieval. It triggers question decomposition and rereading when the final memory is insufficient, enabling the recovery of indirect facts that were prematurely discarded. This design supports non-linear reasoning while preserving the inherent logical flow of document comprehension. To further enhance practicality, we introduce a reinforcement learning framework that enhances length extrapolation capability while dynamically determining the number of rereading passes based on task complexity, thereby flexibly controlling computational overhead. Extensive experiments demonstrate that MemReread consistently outperforms baseline frameworks on long-context reasoning tasks, while maintaining linear time complexity with respect to context length. 

Long-context reasoning tasks have motivated agent-memory approaches that process document chunks sequentially while maintaining a dynamically updated memory, avoiding the quadratic complexity of standard attention. Existing methods often rely on retrieval to recover information discarded during memory overwriting, but remain vulnerable to evidence loss during memory formation and interference from invalid queries. To overcome these limitations, we propose MemReread. Built upon streaming reading, it circumvents intermediate retrieval, triggering question decomposition and targeted rereading when the final memory is insufficient. By recovering indirect evidence that may have been prematurely discarded, MemReread enables non-linear reasoning while preserving the logical flow of document comprehension. We further introduce a reinforcement learning framework that improves length extrapolation and adaptively determines the number of rereading passes according to task complexity, enabling flexible control of computational overhead. Our MemReread-4B outperforms strong baselines on long-context reasoning tasks (e.g., +1.6pp v.s. DeepSeek-V4-Pro on 2WikiMultiHopQA) while maintaining linear time complexity with respect to context length.
\end{abstract}

\section{Introduction}
\label{section:intro}

\begin{figure}[t]
    \centering
    \includegraphics[width=\linewidth]{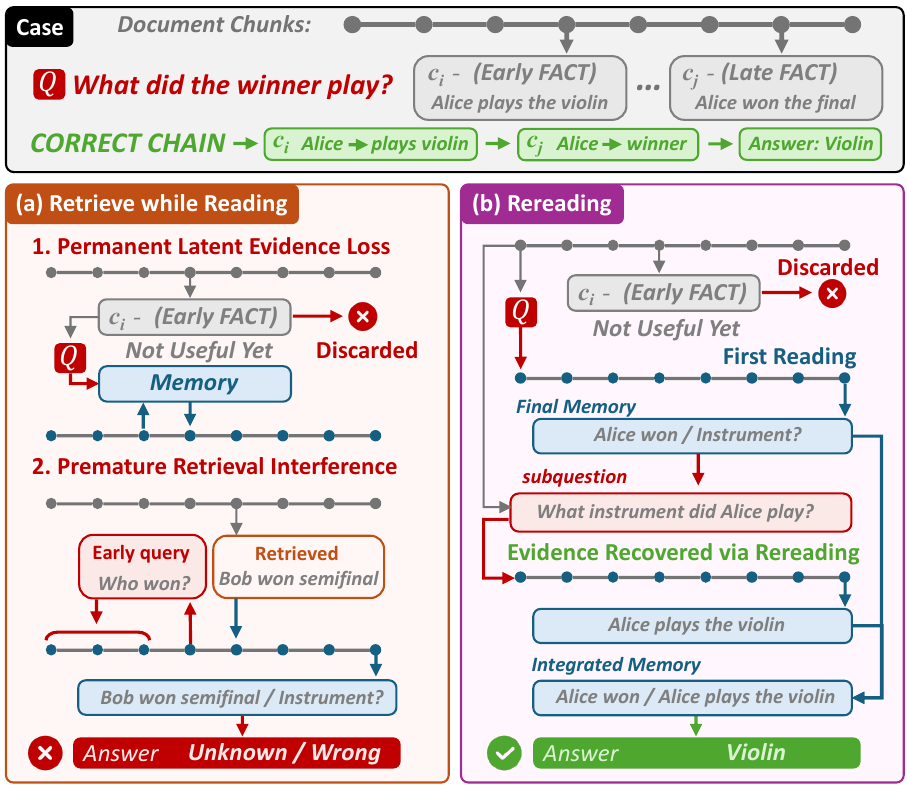}
    \caption{\textbf{Comparison of recalled paradigms.} (a) Limitations of Retrieval-Based Approaches. Latent evidence discarded from chunks, rather than from previous memory, becomes irretrievable. (b) Proposed Mitigation via Rereading with Sub-questions.}

    \label{fig:preliminary}
\end{figure}

Large language models (LLMs) struggle with contexts approaching or exceeding their training window~\cite{hsieh2024ruler, kuratov2024babilong,bai2025longbench}, as attention dilution obscures critical evidence and disrupts coherent reasoning~\cite{hsieh2024found, tang2026revisiting}. Extending context windows through training remains prohibitively expensive because of the quadratic complexity of attention~\cite{li2025training, chen2024longlora, li2026out}.
These limitations have motivated chunk-level frameworks based on retrieval, memory, or their combination. Retrieval-only methods risk semantic fragmentation, query-chunk misalignment, and consequent information loss~\cite{tao2025saki}, whereas the sequential nature of memory-only methods prevents them from revisiting overlooked evidence, limiting non-linear reasoning~\cite{yu2026memagent}. Recent hybrid frameworks combine retrieval-based recall with contextual memory to improve semantic coherence and performance~\cite{shi2026look, wang2026infmem}.

\begin{figure}[t]
    \centering
    \includegraphics[width=\linewidth]{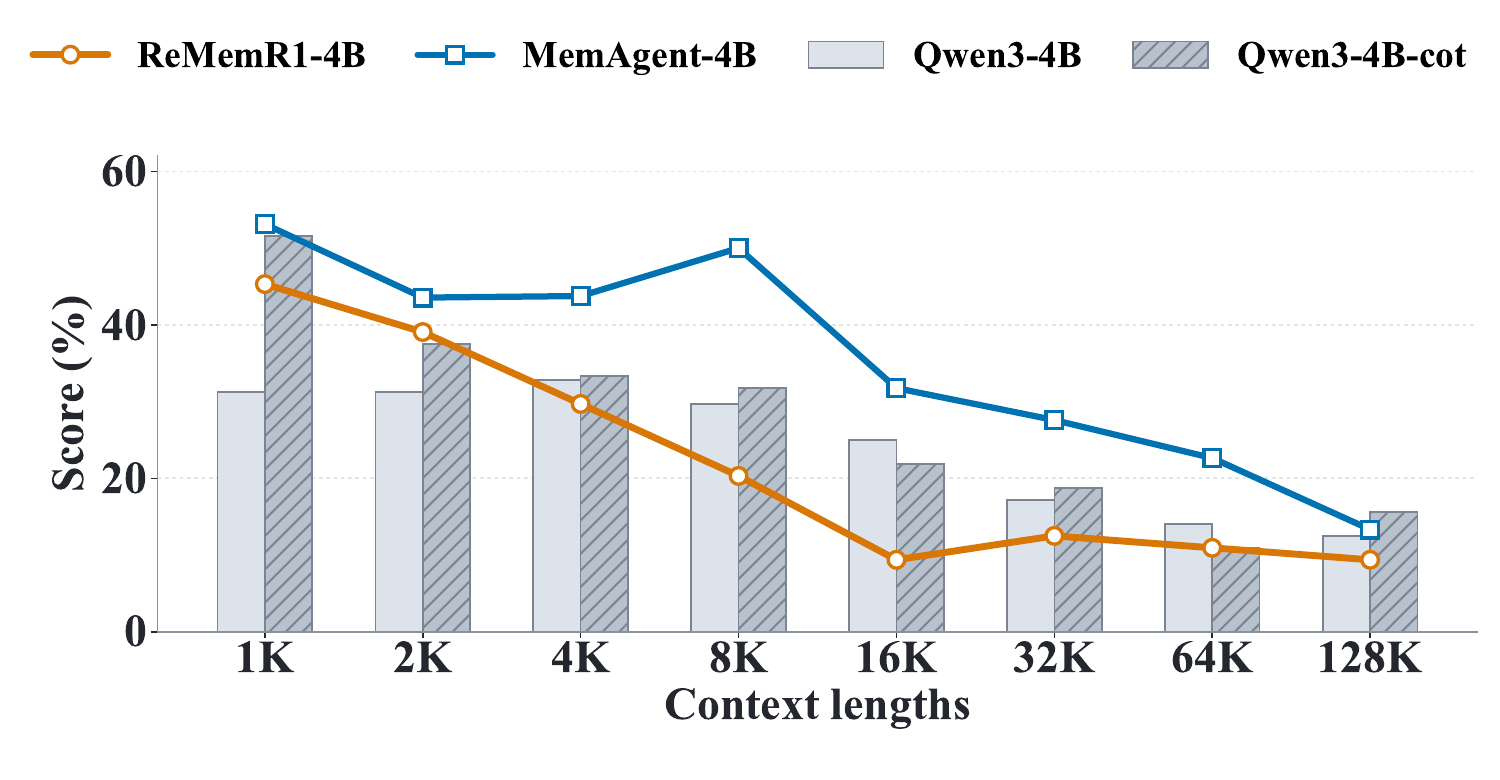}
    \caption{Performance on the Global Reasoning Task.}
    \label{fig:memory_step}
\end{figure}

Despite enabling length-unbounded, non-linear reasoning, these agents remain limited on complex context-dependent tasks for two main reasons:

\begin{itemize}[leftmargin=*, labelindent=0pt]

    \item \textbf{Permanent Latent Evidence Loss.}
Historical-memory recall can recover overwritten information only if it was previously archived. However, latent evidence whose relevance becomes apparent only after later bridge facts are observed is often prematurely discarded as noise. Since such evidence never enters historical memory, it cannot be retrieved later.
% creating an irreversible information loss.

    \item \textbf{Interference from Invalid Queries.}
Retrieval-augmented agents cannot determine whether missing information has been overwritten or has yet to be read. This ambiguity triggers premature queries that retrieve irrelevant content and contaminate memory, progressively diluting critical evidence and degrading long-context reasoning.

\end{itemize}

To address these challenges, we introduce MemReread, as illustrated in Figure~\ref{fig:preliminary}. Guided by accumulated memory, MemReread decomposes the task to identify the highest-priority unresolved sub-question. It then rereads the document for the required evidence, answers the sub-question using the resulting sub-memory, and incorporates the sub-answer into the root memory. This process continues until the root memory is sufficient to answer the original question. 

To control the additional cost of rereading, we design a rereading-count-aware advantage calculation that discourages unnecessary passes while preserving task performance. We further introduce MemReread-Flash, an optional acceleration variant that trades a modest amount of performance for lower inference cost. Extensive experiments show that MemReread outperforms specialized memory agents and DeepSeek-V4-Pro, while MemReread-Flash remains superior to competing memory-agent baselines under a reduced computational budget.

In summary, our contributions are:
\begin{itemize}[leftmargin=*, labelindent=0pt]
    \item We propose MemReread, a memory-agent framework that enables non-linear reasoning through retrieval-free rereading. Regardless of document length, it maintains a context window below 8K tokens while achieving linear time and constant extra space complexity.

    \item We develop a tailored advantage calculation that enables the agent to learn context-aware rereading policies through reinforcement learning and adapt the rereading passes to task complexity.

    \item Experiments demonstrate that MemReread remains effective across model scales and contexts of up to one million tokens.
    % outperforming DeepSeek-V4-Pro on 2WikiMultiHopQA.
\end{itemize}

% \input{sections/2_RelatedWork}
%2.5
\section{Preliminary}
\label{section:preliminary}

\begin{figure}[t]
    \centering
    \includegraphics[width=\linewidth]{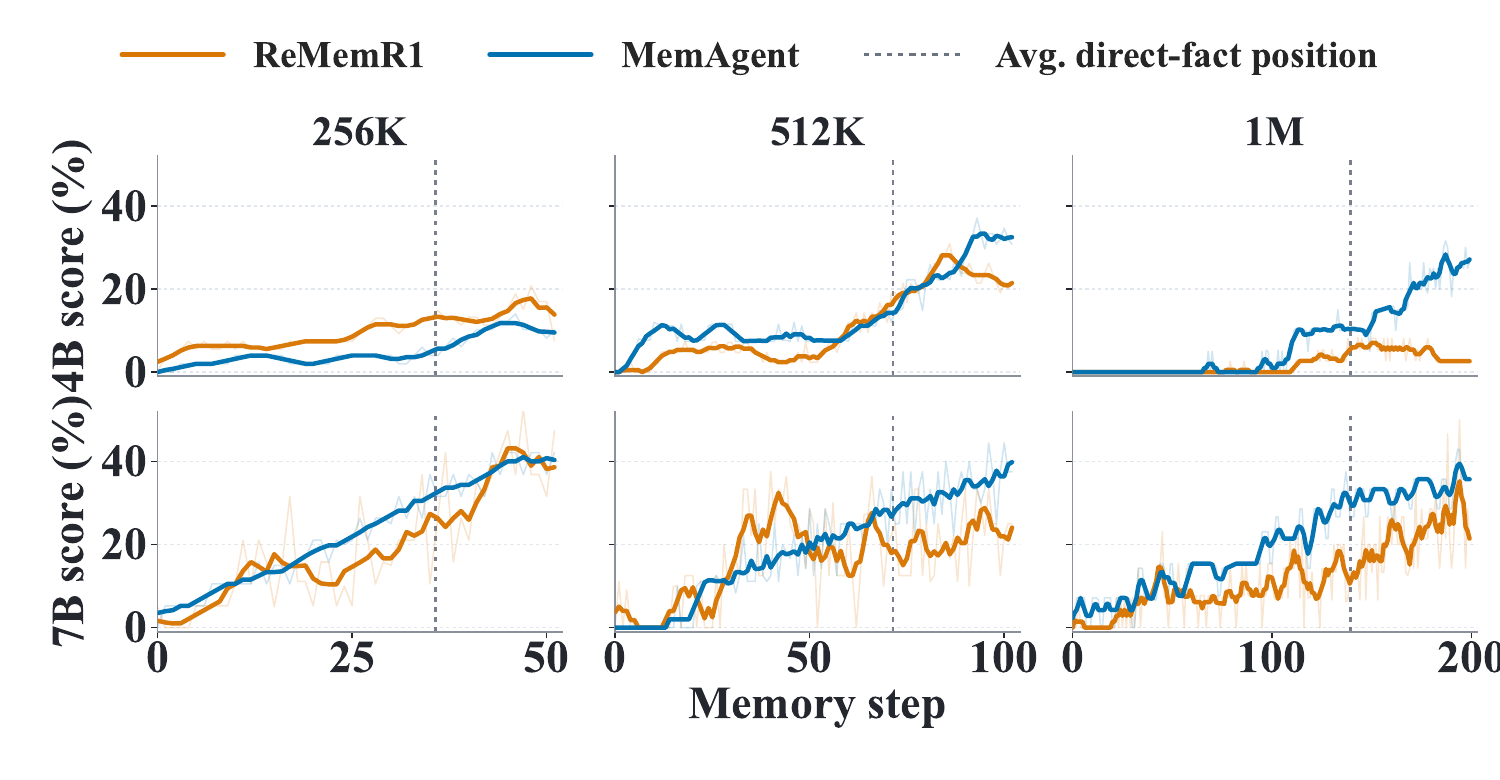}
    \caption{Retrieval-induced degradation.}
    \label{fig:memory_noise}
\end{figure}

First, we review retrieval-augmented memory agents for long-context tasks and analyze their theoretical limitations (Section~\ref{pre_theoretical}). Second, we synthesize a diagnostic dataset to verify retrieval failure modes (Section~\ref{pre_experiment}). Finally, we demonstrate that simple rereadings suffice to mitigate these failures (Section~\ref{pre_solve}).

\subsection{Retrieval-Augmented Memory Agents for Long-Context Reasoning}
\label{pre_theoretical}

We consider the long-context reasoning task, whose sample is given as $(Q, C, A)$, where $Q$ denotes a question, $C$ denotes a long document, and $A$ denotes the answer. 
Memory agents following the memorize-while-reading paradigm~\cite{yu2026memagent, shi2026look, wang2026infmem} partition $C$ into bounded-length chunks $c_0,c_1,\ldots,c_{T-1}$ and process them sequentially. We refer to agents that retrieve information during each chunk-level memory update as Retrieval-Augmented Memory Agents (RA-MemAgents).

The sequential procedure of RA-MemAgents can be cast as a Markov Decision Process (MDP), formulated as $(\mathcal{S}, \mathcal{A}, P, \mathcal{R})$. At each step $t$:
\begin{itemize}[leftmargin=*, labelindent=0pt]
    \item $s_t = (m_t, q_t) \in \mathcal{S}$ denotes the state, where $m_t$ is defined by agent's memory, and $q_t$ is retrieval query generated along with $m_t$.
    \item $a_t \in \mathcal{A}$ denotes the action representing an update to the memory, which is determined by the policy $\pi_{\theta}$ given $Q, c_t, m_t$ and the retrieved content $\mathcal{E}(q_t, \{m_i\}_{i \in [0, t)} | \{c_j\}_{j \in [0, T)})$.

    \item $P(s_{t + 1} | s_t, a_t)$ denotes the transition. Specifically, the transition is defined as:

\begin{equation}
\small
\begin{aligned}
s_{t+1}
&=
\bigl(m_{t+1},q_{t+1}\bigr) =
\pi_{\theta}\Bigl(
    Q,\,
    c_t,\,
    m_t,
\\[-0.3em]
&\mathcal{E}\bigl(
        q_t,\,
        \{m_i\}_{i\in[0,t)}
        \mathrel{\big|}
        \{c_j\}_{j\in[0,T)}
    \bigr)
\Bigr).
\end{aligned}
\label{eq:state_transition}
\end{equation}

    % where $ (m_0,q_0)=(\varnothing,\varnothing)$.
    At the terminal step $T$, $o$ denotes the final answer, derived from $m_T$.
    \item The reward $r_t \in \mathcal{R}$ is defined at each step $t \in [0, T)$, mainly based on the quality of the final answer $o$ after all chunks have been processed.
\end{itemize}

As formalized above, historical-memory retrieval selectively preserves question-relevant facts from $c_i$ in $m_i$, making discarded evidence unrecoverable. Since relevance may become apparent only from later context, useful early facts can be removed before their importance is recognized. Moreover, intermediate retrieval cannot reliably determine whether missing information has been overwritten, remains unread, or is absent, leading to ineffective queries and memory contamination. Chunk-level retrieval, meanwhile, requires storing the entire document, struggles to integrate evidence scattered across chunks, and incurs high peak computation when multiple chunks are processed jointly, limiting its applicability in resource-constrained settings~\cite{shi2026look}.

Given these trade-offs, we focus our comparison and analysis on methods that maintain a context window below 8K tokens at every stage.

\subsection{Retrieval Failure Analysis}
\label{pre_experiment}
\paragraph{\textbf{Global Reasoning Task}}
To empirically examine the limitations of memory retrieval, we construct a diagnostic dataset, termed Global Reasoning, based on RULER-QA~\cite{hsieh2024ruler}. Following RULER's scalable design, we separate evidence from background context to support evaluation from 8K to 1M tokens. The dataset comprises two tasks—statistics and variable tracking—corresponding to the two retrieval limitations illustrated in Figure~\ref{fig:preliminary}. To introduce non-linear evidence dependencies, we place the direct fact—the only fact explicitly related to the question—late in the context, inducing agents to prematurely discard early latent evidence whose relevance becomes apparent only afterward. The background context consists of segmented real-world essays, with full construction details provided in Appendix~\ref{subsection:preliminary_dataset}.

\paragraph{\textbf{Experimental Setups}}
We analyze failure modes at the 4B and 7B scales using Qwen3-4B~\cite{yang2025qwen3} and Qwen2.5-7B-Instruct~\cite{qwen2.5}, respectively. For comparative baselines, we select MemAgent~\cite{yu2026memagent} as a representative of the pure streaming reading paradigm, and ReMemR1~\cite{shi2026look} as a representative of the retrieval-augmented memory agent. We first evaluate on the Global Reasoning Task at the 4B scale. Second, we conduct fine-grained analysis at both 4B and 7B scales, where we treat the recorded memory at each step as the final memory for direct answering, and track accuracy at each memorizing step. More details are provided in Appendix~\ref{subsection:prelinminary_settings}

\paragraph{\textbf{Observation}}

As shown in Figure~\ref{fig:memory_step}, MemAgent outperforms the base model, whereas the retrieval-augmented ReMemR1 exhibits a negative effect. 
Figure~\ref{fig:memory_noise} shows accuracy trends when answering using memory at each step. Both frameworks improve as reading progresses since essential facts are scattered throughout the context. However, MemAgent exhibits a more stable, continuous ascent, while ReMemR1 oscillates severely throughout the process, with its performance even dropping or stalling after encountering the direct fact. We attribute this to retrieval-induced interference that disrupts its ability to count or track information on the Global Reasoning Task. We provide further details in Appendix~\ref{subsection:prelinminary_supplementary}.

 \begin{figure}[t]
    \centering
    \vspace{-0.5em}
    \includegraphics[width=\linewidth]{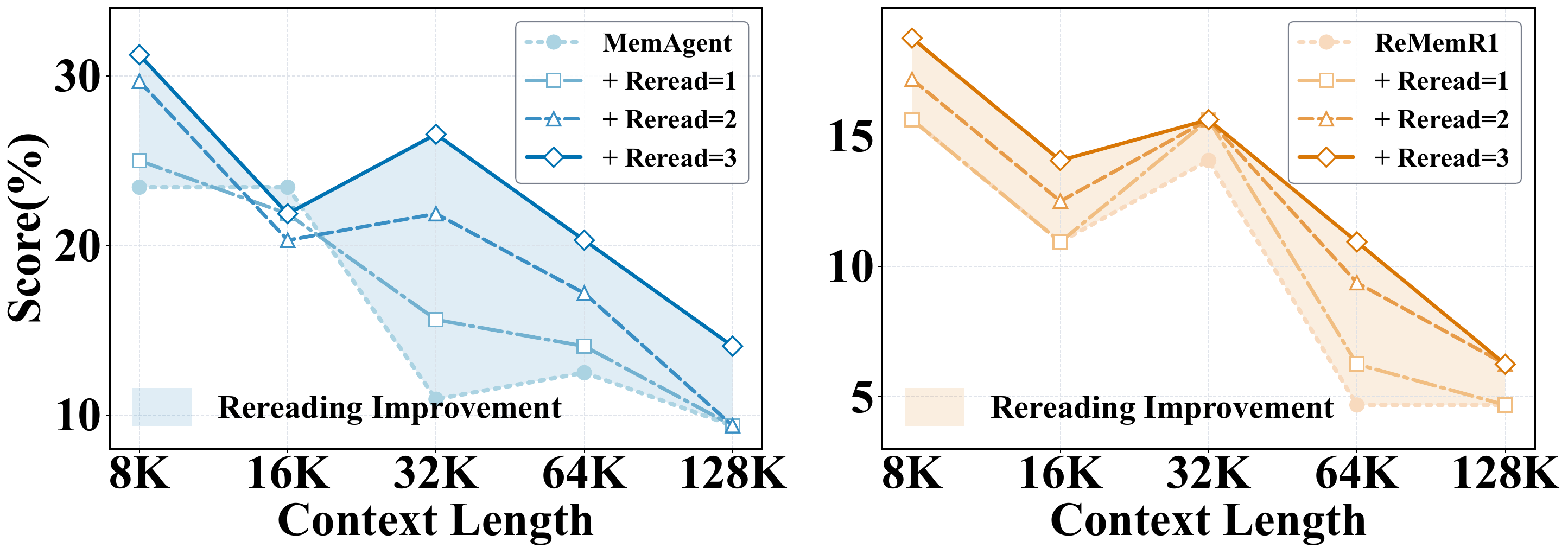}
    \vspace{-1em}
    \caption{Rereading Improvement at 4B Scale.}
    \label{fig:framework_compare}
    \vspace{-0.5em}
\end{figure}

\begin{figure*}[t]
    \centering
    \vspace{-0.5em}
    \includegraphics[width=0.8\linewidth]{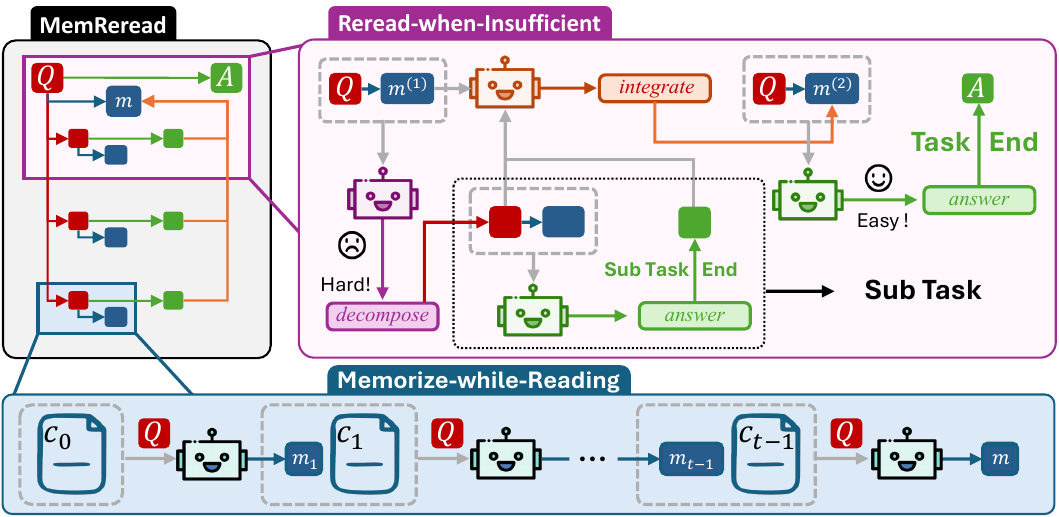}
    % \vspace{-1em}
    \caption{\textbf{Framework of MemReread.} MemReread adopts the streaming memorization paradigm. Upon detecting an information deficit within the terminal memory, it decomposes the initial question, then executes targeted rereading based on the generated sub-questions, and subsequently integrates the derived sub-answers back into the terminal memory.}
    \label{fig:method}
    \vspace{-0.5em}
\end{figure*}

\subsection{Memory Agents with Rereading}
\label{pre_solve}
\paragraph{\textbf{Rereading Mechanism}}

Given these limitations, we maintain the original streaming memorization mechanism during the reading phase. Furthermore, upon completing a reading pass, we prompt the model to identify missing information from the final memory state $m$. If gaps exist, the model generates a sub-question $q$, then performs streaming reading with $q$. It generates the answer $a$ directly from sub-memory upon completion, and finally updates $m$ with the $(q, a)$ pair. This process iterates until $m$ encompasses all necessary information. We detail this process in Section~\ref{section:method_workflow}.

\paragraph{\textbf{Rereading Improvement}}
We apply the rereading mechanism to both MemAgent and ReMemR1 architectures. We set different maximum limits on the number of rereading passes and evaluate on the Global Reasoning Task. As shown in Figure~\ref{fig:framework_compare}, rereading improves performance on both underlying streaming mechanisms. Furthermore, increasing the number of rereading passes yields progressively larger improvements. To circumvent the retrieval-induced interference discussed above, we adopt MemAgent's streaming reading paradigm as the foundation of our framework.

\begin{figure*}[t]
    \centering
    \vspace{-0.5em}
    \includegraphics[width=0.8\linewidth]{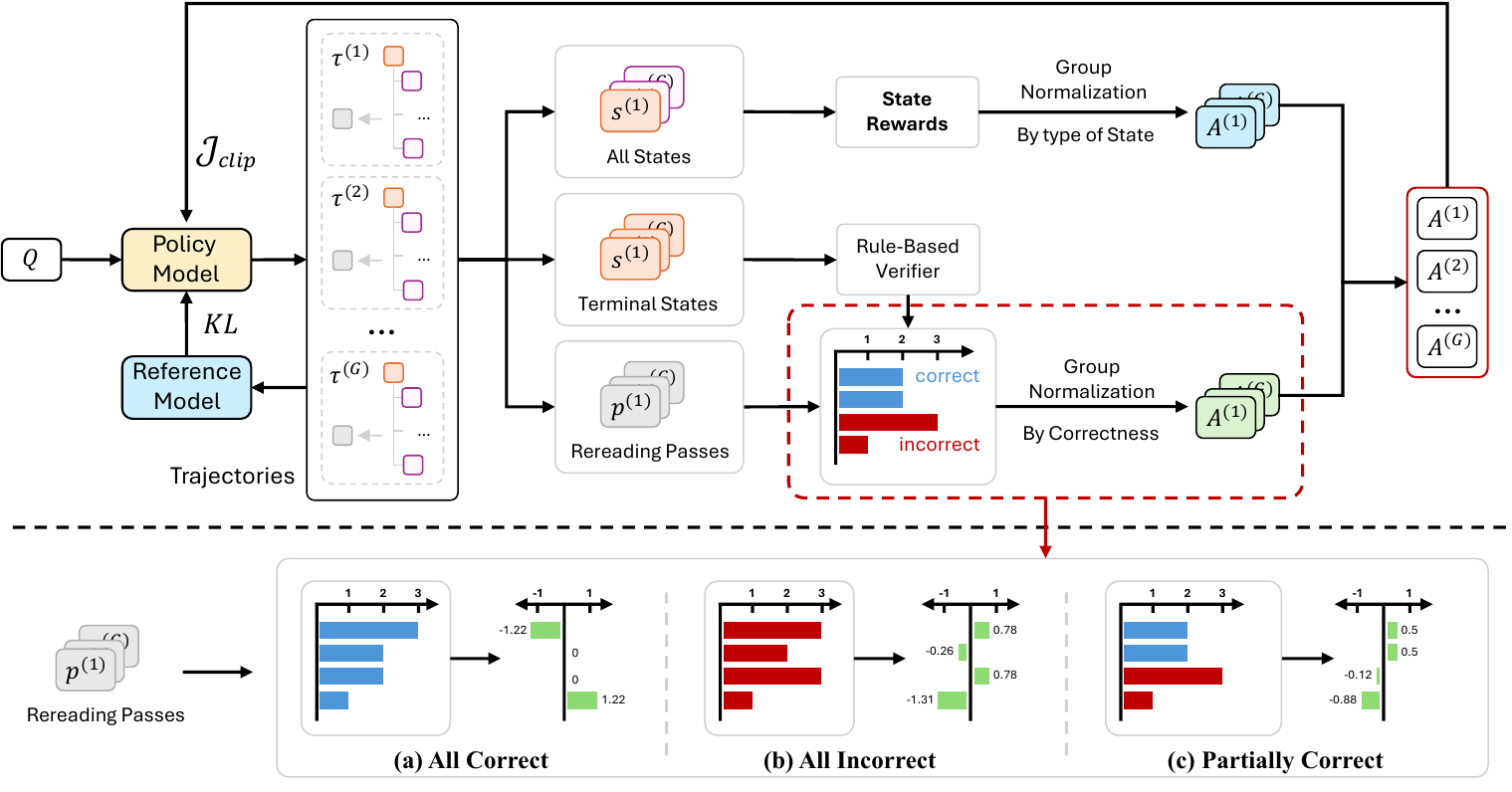}
    \vspace{-1em}
    \caption{\textbf{Overview of the Advantage Design.} The advantage calculation comprises two components. For process supervision, we adopt the memory part of ReMemR1's state rewards. For outcome supervision, we employ a Rereading-Adaptive Outcome Advantage calculation. }
    \label{fig:node_grpo}
    \vspace{-0.5em}
\end{figure*}

\section{Methodology}

In this section, we present MemReread. First, we demonstrate its mechanism(Section~\ref{section:method_workflow}). Then, we propose a training strategy to enhance long-context reasoning performance (Section~\ref{section:method_train}). Finally, we introduce MemReread-Flash, an optional acceleration variant (Section~\ref{section:inference_acceleration}).

\subsection{The MemReread Workflow}
\label{section:method_workflow}
As illustrated in Figure~\ref{fig:method}, MemReread operates in four phases: \textit{Read}, \textit{Decompose}, \textit{Integrate} and \textit{Answer}. Throughout execution, each step requires only one bounded memory that retains context information, which shapes every operational decision. Further details are provided in Appendix~\ref{app:design_logic}.

\paragraph{\textbf{\textit{Read \& Answer}}} 
At these two stages, we adopt the MemAgent~\cite{yu2026memagent} paradigm. During reading, the system processes context chunks sequentially. It maintains a bounded memory buffer throughout ingestion. Upon completing the full context pass, the final memory will be passed to drive subsequent operations. During answer generation, the agent relies exclusively on the memory. It responds without re-accessing the full text.

\paragraph{\textbf{\textit{Decompose}}}
Given the terminal memory from the \textit{Read} stage, the agent assesses whether it contains sufficient evidence to answer the target question. If so, it proceeds to \textit{Answer}; otherwise, it decomposes the question using engineered prompts. Each sub-question must be more specific than the original and advance the reasoning process, thereby guiding the next rereading pass.

\paragraph{\textbf{\textit{Integrate}}}
This stage is triggered exclusively following sub-task resolution. The agent updates the terminal memory solely with the sub-question and its corresponding answer, directly discarding the intermediate sub-memory. Simultaneously, this QA pair is recorded into the decomposition history to prevent the generation of redundant sub-questions in subsequent steps.

\subsection{Training with Rereading-Adaptive GRPO}
\label{section:method_train}

We employ reinforcement learning to enhance length extrapolation. As illustrated in Figure~\ref{fig:node_grpo}, we adopt the separation of process and outcome advantages from ReMemR1, which effectively mitigates training inefficiency caused by sparse rewards. Built upon it, we design a Rereading-Adaptive(ReA) outcome advantage. This objective encourages minimizing the rereading passes without compromising overall performance.

\paragraph{\textbf{Rereading-Adaptive Outcome Advantage}}
We employ rule-based outcome rewards. Trajectories whose final responses match the golden answer receive $R^{(k)} = 1$, others $R^{(k)} = 0$. Differently, our advantage calculation follows two principles:
\begin{itemize}[leftmargin=*, labelindent=0pt]

    \item For rollout groups with the same outcome reward(all correct or all incorrect), we compute GRPO~\cite{shao2024deepseekmath} advantages based on rereading passes. For fully correct groups, we incentivize brevity. Trajectories with fewer rereading passes receive higher advantages. For fully incorrect groups, we encourage additional rereading. Trajectories with more rereading passes receive higher advantages.
    \item For rollout groups with different outcome rewards(partially correct), inspired by DRPO~\cite{li2026drpo}, we assign positive advantages to correct trajectories and negative advantages to incorrect ones. To ensure training stability, the advantages are normalized such that correct trajectories sum to $1$ and incorrect ones sum to $-1$. Within each subset, advantages are further modulated by rereading passes. Among correct trajectories, fewer rereadings yield higher positive advantages. Among the incorrect ones, more rereading yields higher negative advantages.
\end{itemize}

Finally, we denote the outcome advantage as Equation~\ref{eq:advantage}, where $p$ denotes the rereading count.

\begin{equation}
\small
\begin{aligned}
S
  &:= \sum\nolimits_k R^{(k)},\
Z_g:= \sum\nolimits_{\substack{k, R^{(k)}=R^{(g)}}}
      e^{-p^{(k)}},\\[-0.3em]
\hat{A}_{\mathrm{out}}^{(g)}
  &=
  \begin{cases}
  \displaystyle
  (-1)^{\mathbb{I}(S=G)}
  \frac{p^{(g)}-\mu_p}
       {\sigma_p+\epsilon},
  & if \ S\in\{0,G\},\\[0.8em]
  \displaystyle
  (-1)^{R^{(g)}+1}
  \frac{e^{-p^{(g)}}}{Z_g},
  & \text{otherwise}.
  \end{cases}
\end{aligned}
\label{eq:advantage}
\end{equation}

\paragraph{Overall Advantage}

As shown in Figure~\ref{fig:node_grpo}, 
we incorporate process-level advantage, as it enables more fine-grained credit assignment and improves policy optimization~\cite{shi2026look}.
We denote the process advantage as $\hat{A}^{(g)}_{\text{state}, p, t}$, where $p, t$ denote the number of rereading passes and the chunk index, respectively. The complete advantage calculation in Equation~\ref{eq:overall_advantage} comprises two components:

\begin{equation}
\hat{A}^{(g)}_{p, t} = \alpha 
\hat{A}^{(g)}_{\text{out}} +  
(1-\alpha)\hat{A}^{(g)}_{\text{state}, p, t}
\label{eq:overall_advantage}
\end{equation}
where $\alpha$ is a hyperparameter that controls the importance of the outcome advantage. More details are provided in  Appendix~\ref{app:training_adv}.

\begin{table*}[t]
\centering
\vspace{-0.5em}
% \resizebox{\textwidth}{!}{
    \small
    \begin{tabular}{c l | c c c c c c c c | c}
% \begin{tabularx}{\textwidth}{c | C C C C C | C C C C C }
        \toprule
        \multirow{2.5}{*}{\textbf{Scale}} & \multirow{2.5}{*}{\textbf{Framework}} & \multicolumn{8}{c|}{\textbf{\emph{Context Length}}} & \multirow{2.5}{*}{\textbf{Avg.}}  \\
        \cmidrule(lr){3-10}
        & & 8K & 16K & 32K & 64K & 128K & 256K & 512K & 1M  &\\
        \arrayrulecolor{black}\midrule
        
        % \rowcolor{blue!5} \multicolumn{11}{c}{\textit{\textbf{2WikiMultiHopQA (Out-of-Distribution)}}} \\
        1.6T & Deepseek-V4-Pro~\cite{deepseekv4report2026} & \textbf{72.7} & \underline{70.3} & \textbf{70.3} & \textbf{64.8} & \underline{53.1} &  \textbf{57.0} & 40.6 & 25.8 & \underline{56.8} \\
        
        \arrayrulecolor{gray}\midrule
        \multirow{4}{*}{1.7B} 
        & RAG~\cite{lewis2020retrieval} & 30.5 & 22.7 & 18.0 & 14.1 & 14.8 & 17.2 & 23.4 & 15.6 & 19.5 \\
        & FLARE~\cite{jiang2023active} & 36.7 & 28.1 & 34.4 & 27.3 & 22.7 & 17.2 & 19.5 & 16.4 & 25.3 \\
        & MemAgent~\cite{yu2026memagent} & 33.6 & 36.7 & 24.2 & 23.4 & 21.1 & 21.1 & 24.2 & 21.9 & 25.8 \\
        &  ReMemR1~\cite{shi2026look} & 35.2 & 34.4 & 39.1 & 21.9 & 27.3 & 32.0 & 21.1 & 22.7  & 29.2 \\
        &  \textbf{MemReread} & 37.5 & 36.7 & 37.5 & 25.0 & 28.1 & 27.3 & 25.0 & 25.0 & 30.3 \\
        
        \arrayrulecolor{gray}\midrule
         \multirow{4}{*}{4B}& RAG~\cite{lewis2020retrieval} & 43.8 & 41.4 & 36.7 & 44.5 & 31.3 & 32.8 & 29.7 & 28.9 & 36.1 \\
         & FLARE~\cite{jiang2023active} & 54.7 & 52.3 & 41.4 & 41.4 & 32.8 & 35.9 & 40.6 & 36.7 & 42.0 \\
         % & FLARE & 46.9 & 41.4 & 28.9 & 39.1 & 38.3 & 28.1 & 28.1 & 25.0 & 34.5 \\
         &  MemAgent~\cite{yu2026memagent} & 68.0 & 51.6 & 43.8 & 39.1 & 39.8 & 39.1 & 35.2 & 39.8 & 44.6 \\
        &  ReMemR1~\cite{shi2026look} & 64.1 & 54.7 & 44.5 & 37.5 & 41.4 & 49.2 & 37.5 & 41.4  & 46.3 \\
        &  \textbf{MemReread} & \underline{70.3} & \textbf{71.1} & \underline{59.4} & \underline{64.1} & \textbf{54.7} & \underline{55.5} & \textbf{46.9} & \textbf{45.3} & \textbf{58.4} \\
            &  \textbf{MemReread-Flash} & 69.5 & 62.5 & 58.6 & 53.9 & 44.5 & 51.6 & \underline{42.2} & \underline{42.2} & 53.1 \\
        \arrayrulecolor{black}\toprule
% \end{tabularx}
    \end{tabular}
\caption{Results on 2WikiMultiHopQA~\cite{ho2020constructing}. Values are accuracy(\%), rounded to 1 decimal. \textbf{Bold} denotes the best performances, and \underline{underline} denotes the second-best performances.}
% }
\label{tab:main_result}
\end{table*}

\subsection{Inference Acceleration}
\label{section:inference_acceleration}

Since each rereading pass scans the entire document, reading dominates
MemReread's inference latency.
MemReread-Flash reduces this cost by using the
sub-question $q_p$ for BM25 retrieval~\cite{robertson2009probabilistic} and
rereading only the most relevant chunks.
Only the highest-scoring chunks are reread, substantially reducing inference cost while maintaining competitive performance.
 Let
$\operatorname{sim}$ denote the similarity function between the query and chunks. Given a retrieval
ratio $\lambda \in(0,1]$, $\left\lceil \lambda T \right\rceil$ denotes the number of chunks selected at pass
$p$, and $\widetilde C_p$ denotes the resulting chunk sequence.
$\operatorname{TopK}$ selects the highest-scoring chunks while
 restores their original document order:
\begin{equation}
\begin{aligned}
\widetilde C_p
    = 
       \operatorname{TopK}_{c_t\in C}
       \bigl(\operatorname{sim}\bigl(q_p,c_t\bigr),\left\lceil \lambda T \right\rceil\bigr).
\end{aligned}
\label{eq:flash-retrieval}
\end{equation}
The retrieved chunks $\widetilde C_p$ are then sequentially processed to
produce a sub-answer, which is integrated into the root memory. Consistent with MemReread, this process
continues until the memory is sufficient or the rereading
passes limit is reached.

%2.5

\begin{figure}[t]
    \centering
    \includegraphics[width=0.9\linewidth]{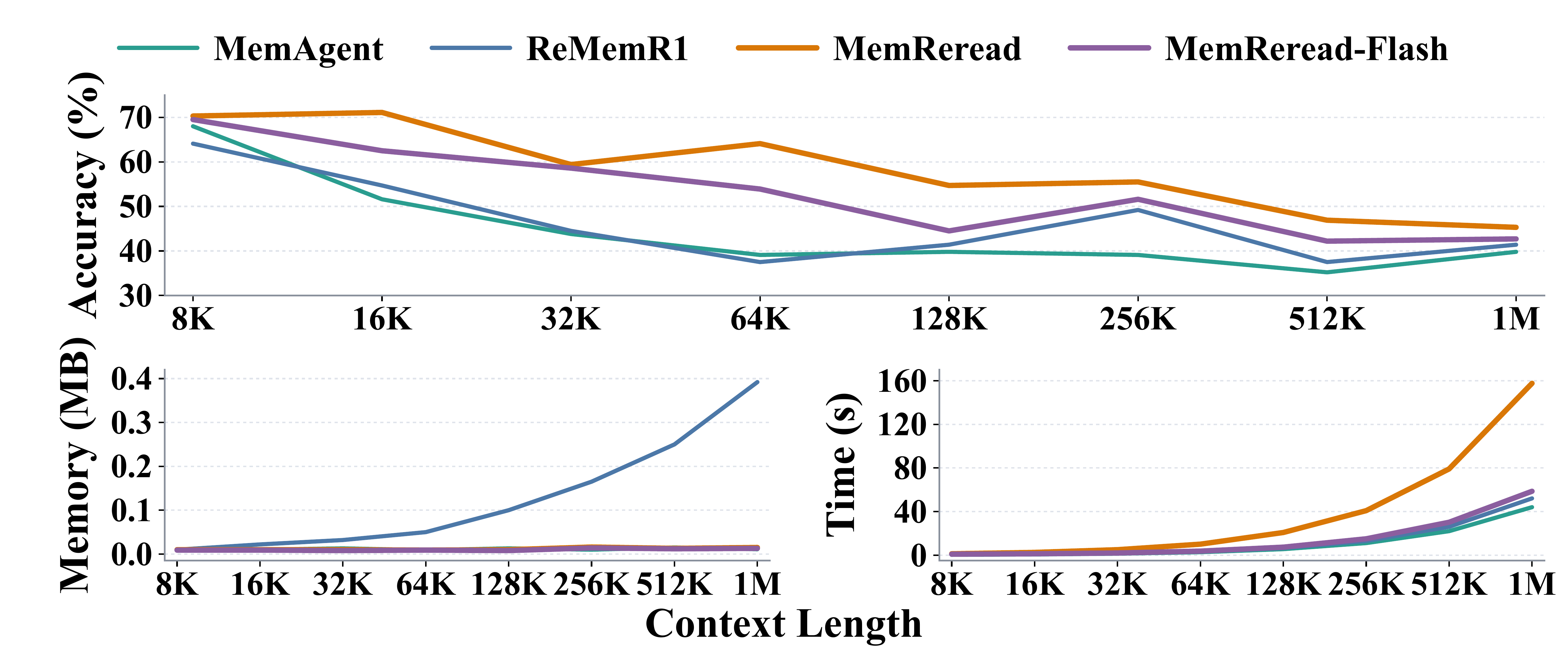}
    \caption{Comparison with memory-augmented agents.}
    \label{fig:time_cost}
\end{figure}

\section{Experiments}
\label{section:experiments}

\begin{table*}[htbp]
    \centering
    \setlength{\tabcolsep}{2pt}
    \small
\resizebox{2\columnwidth}{!}{
    \begin{tabular}{l | c c c c c c c c | c }
        \toprule
        \multirow{2.5}{*}{\textbf{Methods}} & \multicolumn{8}{c|}{\textbf{\emph{Context Length}}} & \multirow{2.5}{*}{\textbf{Avg.}}  \\
        \cmidrule(lr){2-9}
         & 8K & 16K & 32K & 64K & 128K & 256K & 512K & 1M & \\
        \midrule
        \rowcolor{gray!20} \textbf{MemReread(Full)}
         & \textbf{70.3} & \textbf{71.1} & \textbf{59.4} & \textbf{64.1} & \textbf{54.7} & \textbf{55.5} & \textbf{46.9} & \textbf{45.3} & \textbf{58.4}  \\
         \midrule
         ~~~\textit{All Readings $\to$ Retrieval} & 36.7$_{\textcolor{deepred}{\downarrow 47.8 \%}}$ & 28.1$_{\textcolor{deepred}{\downarrow 60.4 \%}}$ & 20.3$_{\textcolor{deepred}{\downarrow 65.8  \%}}$ & 32.8$_{\textcolor{deepred}{\downarrow  48.8 \%}}$ & 21.9$_{\textcolor{deepred}{\downarrow 60.0 \%}}$ & 22.7$_{\textcolor{deepred}{\downarrow 59.1 \%}}$ & 23.4$_{\textcolor{deepred}{\downarrow 50.1 \%}}$ & 19.5$_{\textcolor{deepred}{\downarrow  57.0 \%}}$ & 25.7$_{\textcolor{deepred}{\downarrow 56.0 \%}}$  \\
         ~~~\textit{Rereadings $\to$ Retrieval} & 46.9$_{\textcolor{deepred}{\downarrow 33.3 \%}}$ & 41.4$_{\textcolor{deepred}{\downarrow 41.8 \%}}$ & 28.9$_{\textcolor{deepred}{\downarrow 51.3  \%}}$ & 39.1$_{\textcolor{deepred}{\downarrow  39.0 \%}}$ & 38.3$_{\textcolor{deepred}{\downarrow 30.0 \%}}$ & 28.1$_{\textcolor{deepred}{\downarrow 49.4 \%}}$ & 28.1$_{\textcolor{deepred}{\downarrow 40.1 \%}}$ & 25.0$_{\textcolor{deepred}{\downarrow  44.8 \%}}$ & 34.5$_{\textcolor{deepred}{\downarrow 40.9 \%}}$  \\
         \midrule
         ~~~\textit{w/ \ \ Intermediate Retrieval} & 66.4$_{\textcolor{deepred}{\downarrow \ \ 7.9 \%}}$ & 68.0$_{\textcolor{deepred}{\downarrow \ \ 4.4 \%}}$ & 54.7$_{\textcolor{deepred}{\downarrow \ \ 7.9 \%}}$ & 41.4$_{\textcolor{deepred}{\downarrow 35.4 \%}}$ & 53.9$_{\textcolor{deepred}{\downarrow \ \ 1.5 \%}}$ & 50.8$_{\textcolor{deepred}{\downarrow \ \ 8.5 \%}}$ & 39.1$_{\textcolor{deepred}{\downarrow 16.6 \%}}$ & 40.6$_{\textcolor{deepred}{\downarrow 10.4 \%}}$ & 51.9$_{\textcolor{deepred}{\downarrow 11.1 \%}}$  \\
         ~~~\textit{w/o Decomposition} & 68.8$_{\textcolor{deepred}{\downarrow \ \ 2.1 \%}}$ & 69.5$_{\textcolor{deepred}{\downarrow \ \ 2.3 \%}}$ & 58.6$_{\textcolor{deepred}{\downarrow \ \ 1.3 \%}}$ & 49.2$_{\textcolor{deepred}{\downarrow 23.2 \%}}$ & 54.7$_{\textcolor{deepred}{\downarrow \ \ 0.0 \%}}$ & 48.4$_{\textcolor{deepred}{\downarrow 12.8 \%}}$ & 44.5$_{\textcolor{deepred}{\downarrow \ \ 5.1 \%}}$ & 41.4$_{\textcolor{deepred}{\downarrow \ \ 8.6 \%}}$ & 54.4$_{\textcolor{deepred}{\downarrow \ \ 6.8 \%}}$  \\
         ~~~\textit{w/o Rereading} & 62.5$_{\textcolor{deepred}{\downarrow 1    1.1 \%}}$ & 54.3$_{\textcolor{deepred}{\downarrow 23.6 \%}}$ & 54.7$_{\textcolor{deepred}{\downarrow \ \ 7.9 \%}}$ & 46.1$_{\textcolor{deepred}{\downarrow 28.1 \%}}$ & 43.8$_{\textcolor{deepred}{\downarrow 19.9 \%}}$ & 47.6$_{\textcolor{deepred}{\downarrow 14.2 \%}}$ & 32.8$_{\textcolor{deepred}{\downarrow 30.1 \%}}$ & 35.2$_{\textcolor{deepred}{\downarrow 22.3 \%}}$ & 47.1$_{\textcolor{deepred}{\downarrow 19.3 \%}}$  \\
         ~~~\textit{w/o RL Training} & 60.9$_{\textcolor{deepred}{\downarrow 13.3 \%}}$ & 57.0$_{\textcolor{deepred}{\downarrow 19.8 \%}}$ & 50.8$_{\textcolor{deepred}{\downarrow 14.5 \%}}$ & 46.1$_{\textcolor{deepred}{\downarrow 28.1 \%}}$ & 36.7$_{\textcolor{deepred}{\downarrow 32.9 \%}}$ & 39.8$_{\textcolor{deepred}{\downarrow 28.3 \%}}$ & 32.0$_{\textcolor{deepred}{\downarrow 31.8 \%}}$ & 33.6$_{\textcolor{deepred}{\downarrow 25.8 \%}}$ & 44.6$_{\textcolor{deepred}{\downarrow 23.6 \%}}$ \\
        \bottomrule
    \end{tabular}
}
    \caption{Ablation Study on MemReread. $\to$ denotes replacement. \textcolor{deepred}{$\downarrow$} denotes the performance degradation.}
    
    \label{tab:rereading-value}
\end{table*}

In this section, we report MemReread's long-context performance (Section~\ref{section:exp:main_results}) and analyze its time and space overheads (Section~\ref{subsection:exp:overhead}), performance-efficiency trade-off (Sections~\ref{section:exp:reread_passes}, ~\ref{section:select_lambda}), and component necessity (Sections~\ref{section:adv_reading_vs_retrieval}, \ref{section:exp:decomposition_necessity}). We also examine the effectiveness of ReA-GRPO in dynamically controlling reading passes based on task complexity (Section~\ref{section:exp:node_rl}). Further analyses of scalability, generality, and portability are provided in Appendix~\ref{section:further_analysis}.

\subsection{Setting}
\label{subsection:setting}
\paragraph{Main Datasets} We derive our training corpus from HotpotQA~\cite{yang2018hotpotqa}, extending contexts to approximately 40K tokens via random document augmentation. For length extrapolation evaluation, we employ the out-of-distribution 2WikiMultiHopQA~\cite{ho2020constructing} dataset. Test sequences span lengths from 8K to 1M tokens, measured by the Qwen3~\cite{yang2025qwen3} tokenizer.

\paragraph{Baselines} With Qwen3 as the backbone, we primarily compare against three categories of baselines: (1) Strong LLMs, represented by Deepseek-V4-Pro, (2) RAG-based methods, represented by FLARE, and (3) RL-based memory agents, represented by MemAgent and ReMemR1. Comparisons with additional baselines are provided in Appendix~\ref{app:compare_base}.
\paragraph{Configuration} 
All evaluations are conducted on 4$\times$ NVIDIA A800(80G) GPUs. We set the chunk size to 5K tokens for all memory agents. For MemReread, we set the maximum number of rereading passes $p_{c}=3$. Comprehensive details (including training) are provided in Appendix~\ref{app:design_logic}. 
% Further implementation details are provided in Appendix~\ref{app:training}.

\begin{figure}[t]
    \centering
    \includegraphics[width=0.9\linewidth]{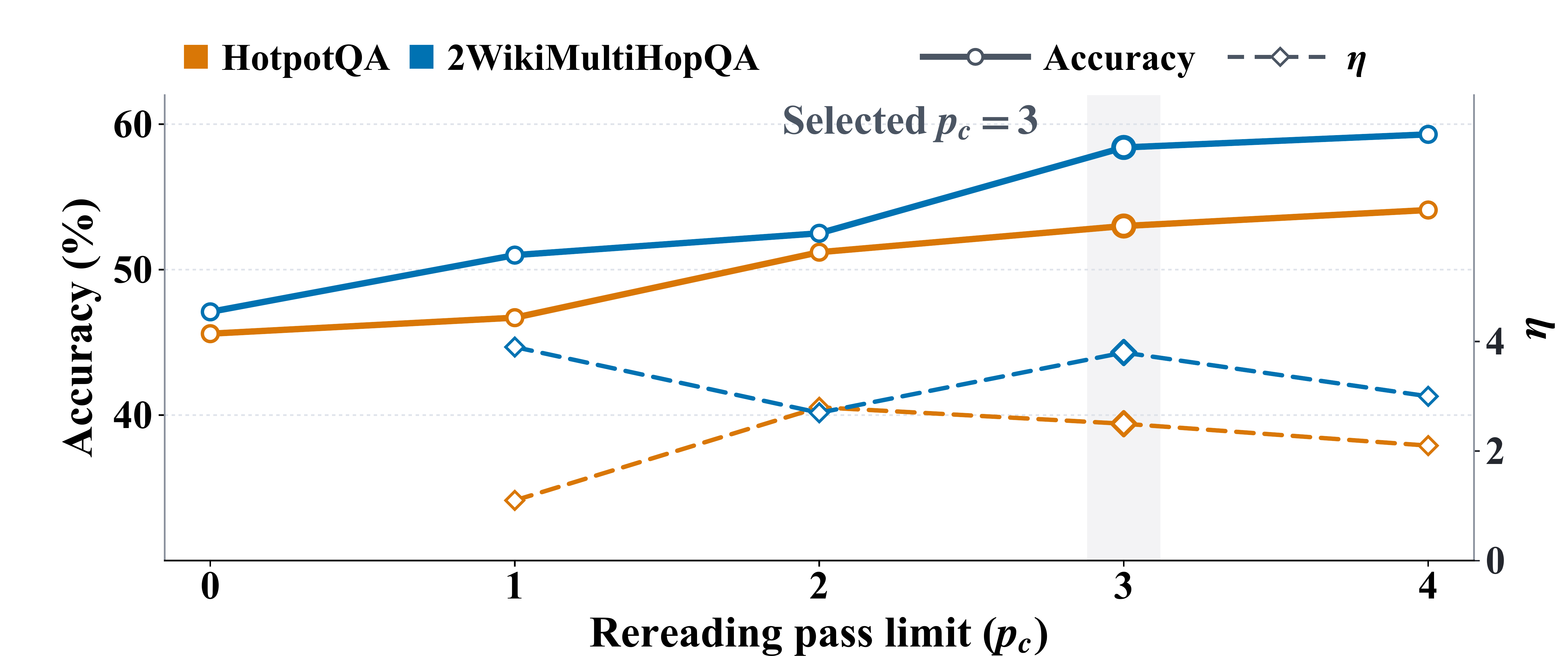}
    \caption{Accuracy and $\eta$ with Different $p_{c}$.}
    \label{fig:eta}
\end{figure}

\subsection{Main Results}
\label{section:exp:main_results}
As shown in Table~\ref{tab:main_result}, MemReread surpasses baseline frameworks across model scales, and RL-based memory agents consistently outperform the RAG-based methods, highlighting the advantage of learned memory management. Despite trading some performance for faster inference, MemReread-Flash still outperforms all competing methods at the same scale. Notably, MemReread-4B exceeds ReMemR1 by up to $12.1$~pp and even outperforms DeepSeek-V4-Pro, suggesting that MemReread learns transferable reasoning strategies rather than merely memorizing training-specific patterns. Additional generalization results are provided in Appendix~\ref{app:more_benchmarks}.

\subsection{Test-Time Overhead Analysis}
\label{subsection:exp:overhead}
\paragraph{Inference Time Overhead}
We compare the average per-sample inference time of MemAgent, ReMemR1, and MemReread across different context lengths. As shown in Figure~\ref{fig:time_cost}, MemReread achieves superior performance on 2WikiMultiHopQA at approximately $3$--$4\times$ the runtime of MemAgent, whereas MemReread-Flash reduces this overhead to only $1.4\times$. The overhead varies with task complexity, as MemReread adaptively determines the number of rereading passes, as detailed in Section~\ref{section:exp:node_rl}. For a context of length $n$ and at most $p_c$ rereading passes, its time complexity is $O(p_c n)$, which remains linear in $n$ because $p_c$ is constant. This linear scaling supports efficient processing of longer contexts, while scalability across model sizes is further discussed in Appendix~\ref{app:scalability}.

\paragraph{Memory Storage Overhead}
% To evaluate the additional space overhead of our rereading design, w
We compare peak stored memory across context lengths for three frameworks: MemAgent, ReMemR1, and MemReread. As shown in Figure~\ref{fig:time_cost}, our rereading mechanism does not require storing historical memory at each step, enabling constant space complexity comparable to MemAgent. In contrast, ReMemR1 stores memory at every step, causing its space usage to scale linearly with context length.

\subsection{Ablation Study}
\label{section:exp:ablation}
\subsubsection{Selection of Maximum Rereading Passes}
\label{section:exp:reread_passes}

Although MemReread adaptively determines rereading passes, it may select excessive passes for challenging samples, leading to unpredictable computational overhead. To constrain this cost, we manually set a maximum rereading limit $p_{c}$ for MemReread. To determine the optimal limit, we evaluate performance with $p_{c} \in \{0, 1, 2, 3, 4\}$ on HotpotQA and 2WikiMultiHopQA. Notably, our method reduces to the standard MemAgent framework when $p_{c}=0$. As shown in Figure~\ref{fig:eta}, increasing $p_{c}$ consistently improves performance across nearly all context lengths on both datasets.

To quantify the performance-efficiency trade-off of increasing $p_{c}$, we define the average per-pass performance gain over the baseline ($p_{c}=0$) as $\eta_{p_{c}=k} = (\mathrm{Avg}_{p_{c}=k} - \mathrm{Avg}_{p_{c}=0}) / k$ for $k \in \{1, 2, 3, 4\}$. We observe diminishing marginal returns in overall performance at $p_{c}=4$, with the average gain per rereading pass declining. Meanwhile, each unit increment of $p_{c}$ increases the upper bound of inference time by the cost of one full streaming reading pass. Therefore, we select $p_{c}=3$ for actual inference tasks. Although larger $p_{c}$ may yield further gains, we do not evaluate them due to the prohibitive inference time cost. We demonstrate in Section~\ref{section:exp:node_rl} that $p_{c}=3$ suffices across representative long-context benchmarks.

\subsubsection{Selection of the Partial-Rereading Ratio}
\label{section:select_lambda}
Reducing the rereading coverage ratio $\lambda$ accelerates inference but increases the risk of missing critical evidence, potentially degrading performance. We examine this trade-off by varying $\lambda$ and reporting the corresponding performance loss and inference speedup in Table~\ref{tab:rereading_ratio}. The results show that $\lambda=1/8$ limits the performance loss to less than $10\%$  while achieving a nearly $2.5\times$ speedup. Based on this favorable balance between effectiveness and efficiency, we adopt $\lambda=1/8$ as the default setting for MemReread-Flash.

\subsubsection{Advantages of Reading over Retrieval}
\label{section:adv_reading_vs_retrieval}
To assess the advantage of rereading over retrieval, we evaluate two ablated variants on the main test set: one replaces all reading passes with retrieval, while the other retains the initial reading pass but replaces subsequent rereading with retrieval. As shown in Table~\ref{tab:rereading-value}, these variants incur performance drops of $56.0\%$ and $40.9\%$, respectively. The larger degradation of the fully retrieval-based variant highlights the importance of initial reading, while the remaining substantial loss confirms the advantage of rereading over retrieval.

\subsubsection{Necessity of Question Decomposition}
\label{section:exp:decomposition_necessity}

To investigate the necessity of question decomposition, we remove the Decomposition component and its associated Integration component, and have the model linearly process four concatenated copies of the document. As shown in Table~\ref{tab:rereading-value}, this variant incurs a $6.8\%$ performance drop. This result demonstrates that simply allocating more computation to repeated reading is insufficient; effective rereading requires decomposition to target unresolved evidence and integration to reason over accumulated evidence without continually expanding memory.

\begin{table}[t]
\centering
\small
\begin{tabular}{lccccc}
\toprule
\multirow{2.5}{*}{\textbf{Metric}}
& \multicolumn{5}{c}{\textbf{Rereading Coverage Ratio $\lambda$}} \\
\cmidrule(lr){2-6}
& $1/16$ & $1/8$ & $1/4$ & $1/2$\\
\midrule
Accuracy (\%)
& 48.7  & 53.1 & 53.4 & 55.1  \\
Accuracy Drop $\downarrow$ (\%)
& 16.6 & 9.1 & 8.6 & 5.7  \\
Inference Speedup ($\times$) 
& 2.94 & 2.42 & 1.88 & 1.26 \\
\bottomrule
\end{tabular}

\caption{
Effect of the rereading coverage ratio $\lambda$ on accuracy and inference
efficiency, which are measured relative to
full rereading ($\lambda=1$).
}
\label{tab:rereading_ratio}
\end{table}

\subsubsection{Effectiveness of ReA-GRPO}
\label{section:exp:node_rl}

\begin{figure}[t]
    \centering
    \includegraphics[width=\linewidth]{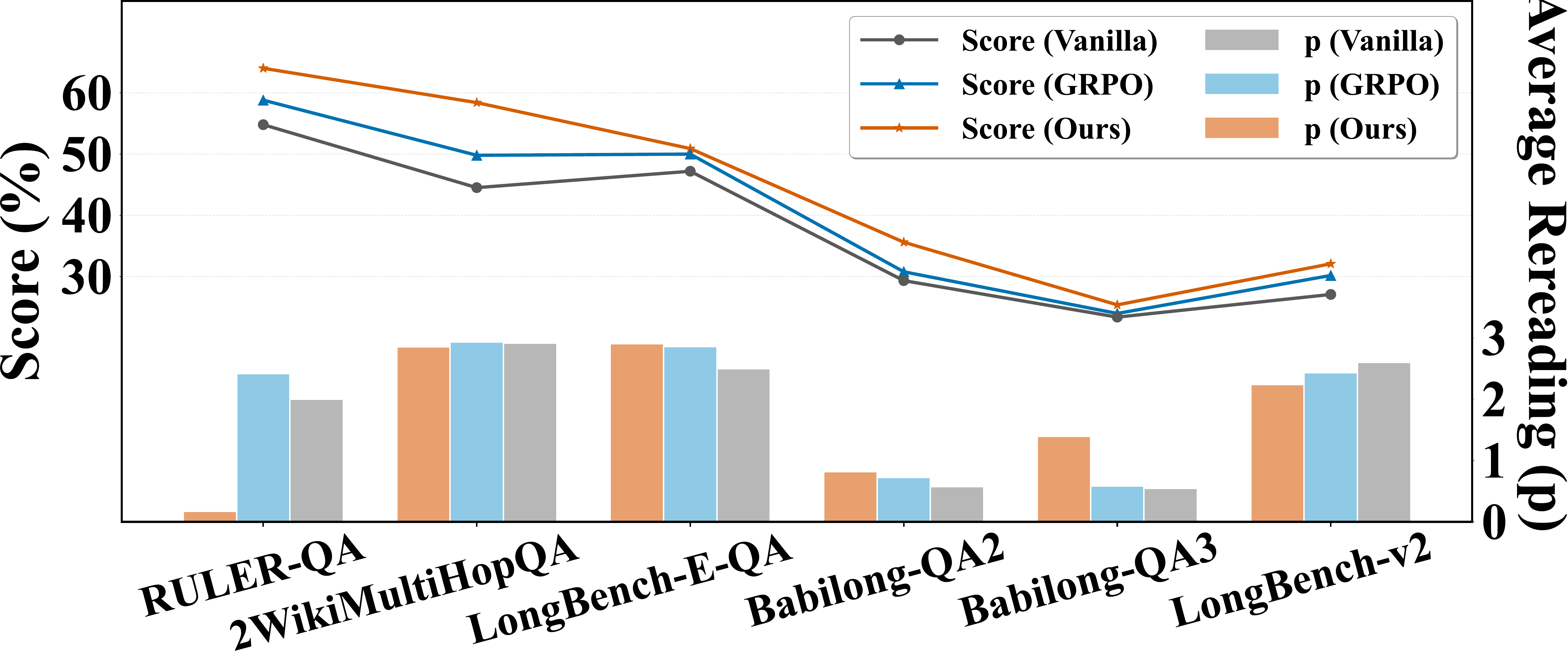}
    \caption{\textbf{Effect of Rereading-Adaptive GRPO.} Vanilla indicates results without RL training.}
    \label{fig:rl_comparison_benchmark}
\end{figure}

In addition to 2WikiMultiHopQA, we evaluate reasoning subtasks from four benchmarks: RULER-QA~\cite{hsieh2024ruler}, LongBench-E-QA~\cite{bai2024longbench}, BabiLong~\cite{kuratov2024babilong}, and LongBench-v2~\cite{bai2025longbench}. With $p_c=3$ for both methods, we compare task performance and the average number of reading passes. As shown in Figure~\ref{fig:rl_comparison_benchmark}, ReA-GRPO consistently outperforms the GRPO baseline while adapting MemReread's reading budget to task demands. Notably, it performs almost no rereading on the largely linear RULER-QA tasks, while its rereading counts on BabiLong QA2 (one-hop) and QA3 (two-hop) closely track their required reasoning hops. Moreover, MemReread uses more passes on LongBench-E-QA than on LongBench-v2 despite achieving higher accuracy, as the former emphasizes multi-hop integration of dispersed evidence, whereas the latter relies more on noise robustness, fine-grained extraction, and internal reasoning. These results demonstrate MemReread's strong task adaptivity. Training curves are provided in Appendix~\ref{app:training}.

%2
% \input{sections/6_Ablation}
%1.5
% \input{sections/7_5_Further_Analysis}

\section{Related Work}
\label{app:related_work}

The quadratic cost of self-attention has motivated extensive research into efficient sequence modeling architectures~\cite{gu2023mamba, de2024griffin, lieber2024jamba, munkhdalai2024leave, liu1889ring, xiao2023efficient, chen2306extending, peng2023yarn, ding2024longrope, su2024roformer}. While these architectures provide foundations for extended-context reasoning, they trade contextual fidelity for computational efficiency and remain limited on unbounded documents~\cite{hsieh2024ruler}. External-memory LLM agents offer an alternative~\cite{hu2025memory}, typically retrieving information evicted during context rotation or memory overwriting~\cite{chhikara2025mem0, xu2025amem, li2025memos, wei2025evo, liang2025sage}. For ultra-long documents, memorize-while-reading agents sequentially process bounded chunks while maintaining a fixed-capacity memory~\cite{xu2025mem, li2025memos, shi2026memocr, yu2026memagent}, with recent systems incorporating retrieval to support cross-chunk reasoning~\cite{shi2026look, wang2026infmem}. More recently, RL-based memory agents have replaced heuristic management with learned policies that adapt memory updates, retrieval, and allocation to long-context reasoning demands~\cite{wen2026reinforcement, yu2026memagent, shi2026look, wang2026infmem}.

\section{Conclusion}

In this work, we propose MemReread, a framework that enhances the streaming reading paradigm by recovering overlooked facts through an adaptive rereading mechanism. To optimize it, we introduce Rereading-Adaptive GRPO (ReA-GRPO), a context-aware reinforcement learning strategy that adaptively modulates rereading passes based on task complexity. Empirical results demonstrate superior performance over baselines, and ReA-GRPO equips it with task-adaptive rereading capabilities. Ablation studies further confirm that increasing the maximum number of rereading passes consistently enhances performance. We hope this work provides valuable insights for the broader academic and engineering community, advancing agentic long-context reasoning capabilities.

% \section{Limitation and Future Work}

\section*{Limitations}
\label{section:limitation}

We identify three primary limitations of this work:

\begin{itemize}[leftmargin=*, labelindent=0pt]

    \item \textbf{Task Generalization.}
    Our evaluation primarily focuses on long-context reasoning. Whether MemReread generalizes to other long-context tasks, such as code completion, text summarization, and long-form generation, remains to be empirically validated.

    \item \textbf{Efficiency--Performance Trade-off.}
    The additional rereading passes of MemReread increase inference latency compared with single-pass streaming methods. Although MemReread-Flash reduces this overhead through selective rereading, its acceleration comes at a non-negligible performance cost. Achieving a better balance between efficiency and effectiveness therefore remains an important direction for future work.

    \item \textbf{Dependence on Backbone Capabilities.}
    MemReread remains bounded by the intrinsic reasoning capabilities of its backbone model. Even when the necessary evidence is recovered and no cross-chunk logical disconnect exists, errors may still occur during decomposition, evidence integration, or answer generation.

\end{itemize}

\bibliography{reference}

\appendix
\begin{table*}[htbp]
    \centering
    \vspace{-0.5em}
    \small
    \resizebox{\textwidth}{!}{%
        \begin{tabular}{l l >{\raggedright\arraybackslash}p{0.3\textwidth} |>{\raggedright\arraybackslash}p{0.3\textwidth}| >{\raggedright\arraybackslash}p{0.3\textwidth}}
            \toprule
            \textbf{Category} & \textbf{Subset} & \textbf{Entity Names} & \textbf{Entity Substitution} & \textbf{Event Substitution} \\
            \midrule
            
            \textbf{Cities} & \textbf{Statistics} & 
            "City\_A", "City\_B", "City\_C", "City\_D", "City\_E", "Nova\_Prime", "Zion", "Matrix" & 
            \multirow{2}{0.3\textwidth}{\raggedright "Code-Alpha", "Code-Beta", "Code-Gamma", "Omega-Protocol", "Sector-X", "Phantom-9", "Alias-77", "Echo-Base", "Node-Zero", "Cluster-V"} & 
            "Operation 77-B", "Protocol X-9", "Class-IV atmospheric disturbance"... \\
            
            \cmidrule(lr){1-3} \cmidrule(lr){5-5} 
            
            \textbf{Variables} & \textbf{Variable Tracking} & 
            "sys\_timeout", "db\_port", "cache\_size", "max\_retries", "log\_level", "worker\_count" & 
            & 
            "core engine parameter", "registry key 0x0FA", "subsystem coefficient"... \\
            \bottomrule
        \end{tabular}%
    }
    \caption{Categories, roles, and example names used in the Global Reasoning Task construction.}
    \label{tab:categories_names_combined}
\end{table*}

\begin{table*}[htbp]
\centering
\vspace{-0.5em}
\small
\resizebox{\textwidth}{!}{%
\begin{tabular}{l l | c c c c c c c c c c c | c}
    \toprule
    \multirow{2.5}{*}{\textbf{SubTask}} & \multirow{2.5}{*}{\textbf{Background Source}} & \multicolumn{11}{c|}{\textbf{\emph{Context Length}}} & \multirow{2.5}{*}{\textbf{Total}} \\
    \cmidrule(lr){3-13}
    & & 1K & 2K & 4K & 8K & 16K & 32K & 64K & 128K & 256K & 512K & 1M & \\
    \midrule
    statistics & Paul Graham Essays & 64 & 64 & 64 & 64 & 64 & 64 & 64 & 64 & 64 & 64 & 64 & 704 \\
    variable-tracking & Paul Graham Essays & 64 & 64 & 64 & 64 & 64 & 64 & 64 & 64 & 64 & 64 & 64 & 704 \\
    \bottomrule
\end{tabular}%
}
\caption{Distribution statistics across different context lengths of the Global Reasoning Task.}
\label{appdix:tab:preliminary_source}
\end{table*}

\begin{figure*}[htbp]
    \centering
    \includegraphics[width=\linewidth]{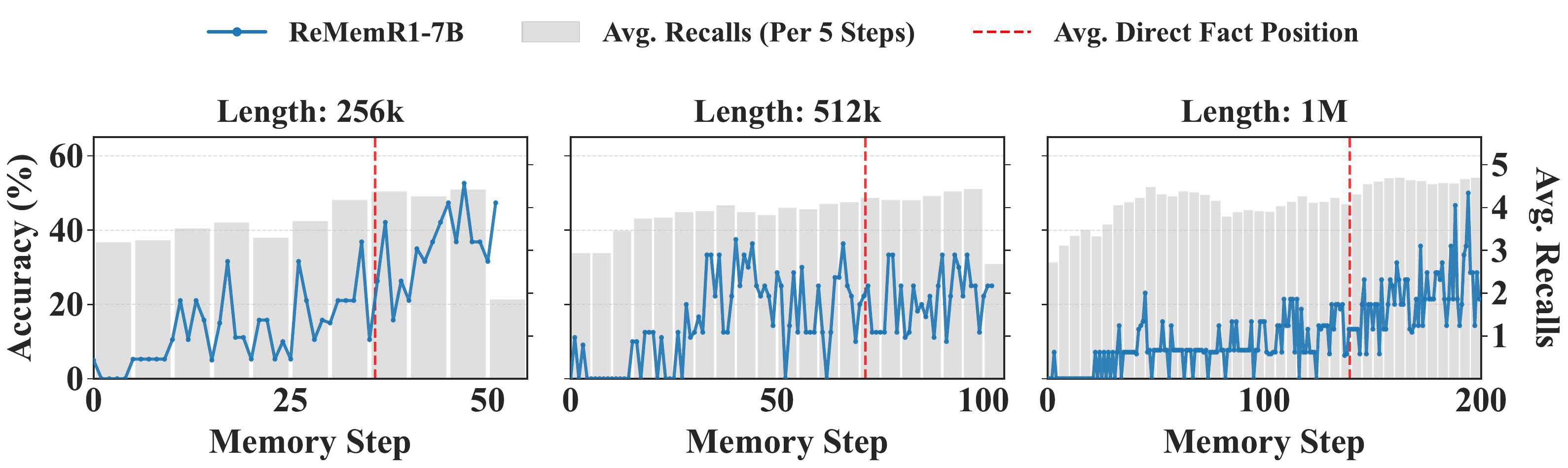}
    \caption{Line plot of performance annotated with retrieval counts.}
    \label{fig:7B_global_recall}
\end{figure*}

\section{Preliminary Study Details}
\label{app:preliminary_details}

\subsection{Global Reasoning Task}
\label{subsection:preliminary_dataset}

To validate the retrieval failure modes identified in our analysis, we construct a dataset targeting the two failure patterns illustrated in Figure~\ref{fig:preliminary}, using RULER-QA~\cite{hsieh2024ruler} as the source dataset. 

Our data construction pipeline comprises two stages: (a) generating multi-hop short-context facts, and (b) inserting these facts relatively uniformly into the background context. For short-context facts, we draw inspiration from RULER~\cite{hsieh2024ruler} to synthesize two task categories: \textbf{Statistics} and \textbf{Variable Tracking}. These tasks evaluate the model's capacity for global latent information aggregation and long-range latent dependency tracking, respectively.

\paragraph{(a) Facts Construction.}
We define \textit{direct facts} as those directly related to the original multi-hop question, and \textit{indirect facts} as those not directly related to the question but crucial for deriving the answer. Instead of directly adopting the key facts from RULER, we employ custom-designed fact structures. We illustrate this design with two representative examples:

\begin{itemize}[leftmargin=*, labelindent=0pt]
    \item \textbf{Statistics:} The \textit{indirect fact} states \textit{``Event X occurred at Location M.''}. The \textit{direct fact} specifies \textit{``Location M is in City A; X is a type of M-event; Location N is not in City A.''}.  A \textit{distractor} notes \textit{``Event X occurred at Location N.''}. The question asks \textit{``How many M-events occurred in City A?''}. Upon encountering the direct fact inserted in the latter half of the context, the model must revisit earlier content to distinguish M-events from non-M-events and perform counting. This evaluates the model's ability to retain implicit information.
    \item \textbf{Variable Tracking:} The \textit{indirect fact} is formatted as \textit{``[System Log Seq N] 'A' of 'M' is updated to 'xxx'.''}. The \textit{direct fact} is formatted as \textit{``M is an alias for X. B is described as A.''}. The \textit{distractor} notes \textit{``[System Log Seq N] 'A' of 'Y' is updated to 'xxx'.''}. The question asks \textit{``According to the system log, what is the final B of X? The final value is determined by the largest Seq N.''}. Upon encountering the direct fact in the latter half of the context, the model must revisit all preceding facts to identify the entry with the Seq N.
\end{itemize}

Note that entity names (e.g., ``City,'' ``Location,'' ``Value of X'') are not unique across samples. To construct non-linear contexts, we perform implicit entity substitution, as in the above two cases. To maintain semantic coherence, we design task-related entities and events for substitution. All entities and substitutions used are listed as detailed in Table~\ref{tab:categories_names_combined}. To ensure moderate task difficulty, we randomly sample an integer from 3 to 10 as the number of facts for each sample.

% 需要说明的是，为避免任务过难或过容易，我们随机选择3~10之间的整数作为线索数量。

\paragraph{(b) Context Padding.}
Upon generating the multi-hop short-context facts, we randomly select a starting point from the background corpus and continuously extract sentences to populate the context until the target length is reached. Subsequently, \textit{indirect facts} are inserted relatively uniformly at random positions throughout the context, whereas the \textit{direct fact} is constrained to the latter half (randomly inserted between positions 0.5 and 0.9). This placement prevents the model from attending to direct facts too early, which would otherwise render all subsequent indirect facts directly relevant to the query and memory state, thereby reducing the task to linear reasoning.

To further illustrate our task design, we provide two data samples in Table~\ref{tab:preliminary_cases}. The task distribution and sequence length statistics are summarized in Table~\ref{appdix:tab:preliminary_source}. Importantly, our task design is geared towards more common patterns found in reasoning tasks, rather than edge cases. We provide a concrete example from another dataset in Table~\ref{tab:2wiki_non_linear_case}.

\subsection{Setting}
\label{subsection:prelinminary_settings}
We conduct preliminary experiments at 4B and 7B model scales. For the 4B setting, since pretrained weights are not publicly available in the open-source community, we adopt the native training frameworks of MemAgent and ReMemR1 with their original configurations, and select the best checkpoints based on validation set performance. For the 7B setting, while ReMemR1-7B weights are publicly released, MemAgent-7B weights are not. Due to computational constraints, we do not reproduce MemAgent-7B from scratch; instead, we directly employ the released ReMemR1-7B weights as the backbone model and evaluate them within the MemAgent framework.

\subsection{Supplementary Results}
\label{subsection:prelinminary_supplementary}

To further verify that it is retrieval that accounts for performance anomalies, we group memory steps into bins of 5 and count the number of retrieval operations within each bin, as shown in Figure~\ref{fig:7B_global_recall}. Although we restrict the number of facts per sample in the Global Reasoning Task to no more than 10, the observed retrieval count substantially exceeds this limit. This suggests that a large portion of retrievals are redundant, failing to provide novel information.
A closer inspection of multiple cases reveals that MemAgent maximizes the preservation of indirect facts during reading, whereas ReMemR1 suffers from premature latent evidence discarding and interference from ineffective retrieval, as shown in Table~\ref{tab:memagent_vs_rememr1_on_gr}.

% \definecolor{green}{HTML}{159447}
% \definecolor{red}{HTML}{B02A37}
% \definecolor{cyan}{HTML}{0057A8}

\begin{table*}[htbp]
    \centering
    \setlength{\tabcolsep}{6pt}
    \renewcommand{\arraystretch}{1.25}
    \begin{tabular}{p{0.95\linewidth}}
    \toprule
    \multicolumn{1}{c}{\textbf{Sample 1: Statistics}} \\
    \hline
    \textbf{Question}: How many distinct magic anomalies were registered in the facility in City\_A? \\
    \midrule
    \textbf{Answer}: 3 \\
    \midrule
    \textbf{Document}: {\footnotesize 
    \textcolor{cyan!60!black}{The facility in Sector-X registered a Category-B logical paradox of type f324d118-ab7c-416f-b3e9-c0404935e14e. \textit{(indirect)}} 
    \textcolor{black}{...(context)...} 
    \textcolor{red!75!black}{The facility in Omega-Protocol registered a Category-B logical paradox of type b27e2d6e-04b6-4727-b81b-9369da5ae7ea. \textit{(distractor)}} 
    \textcolor{black}{...(context)...} 
    \textcolor{cyan!60!black}{The facility in Sector-X registered a Category-B logical paradox of type 1327f6d7-bbb5-4c1f-af49-a78bad83b9d3. \textit{(indirect)}} 
    \textcolor{black}{...(context)...} 
    \textcolor{red!75!black}{The facility in Omega-Protocol registered a Category-B logical paradox of type d9769394-ad5a-4403-9fdb-d7c1348f2f53. \textit{(distractor)}} 
    \textcolor{black}{...(context)...} 
    \textcolor{cyan!60!black}{The facility in Sector-X registered a Category-B logical paradox of type 1aca0b49-b0a7-4f03-950b-0a5b2aecc0be. \textit{(indirect)}} 
    \textcolor{black}{...(context)...} 
    \textcolor{red!75!black}{The facility in Omega-Protocol registered a Category-B logical paradox of type d9769394-ad5a-4403-9fdb-d7c134234523. \textit{(distractor)}} 
    \textcolor{black}{...(context)...} 
    \textcolor{green!50!black}{Note for all personnel: The facility formally designated as Sector-X is physically located in City\_A, and a `Category-B logical paradox' is the official designation for a magic anomaly.  Omega-Proto is not in City\_A. \textit{(direct)}} 
    \textcolor{black}{...(context)...} 
    \textcolor{red!75!black}{The facility in Omega-Protocol registered a Category-B logical paradox of type 1644c6b2-2252-44e7-bdca-33265229bc3a. \textit{(distractor)}}} \\
    \hline
    \multicolumn{1}{c}{\textbf{Sample 2: Variable Tracking}} \\
    \hline
    \textbf{Question}: According to the system logs, what is the final configuration value of `log\_level' (indicated by the highest log sequence number)? \\
    \midrule
    \textbf{Answer}: 1434 \\
    \midrule
    \textbf{Document}: {\footnotesize 
    \textcolor{cyan!60!black}{[System Log Seq 000] The thread pool minimum size `Echo-Base' is initially set to `9673'. \textit{(indirect)}} 
    \textcolor{black}{...(context)...}
    \textcolor{cyan!60!black}{[System Log Seq 003] ... \textit{(indirect)}} 
    \textcolor{black}{...(context)...}
    \textcolor{red!75!black}{[System Log Seq 003] The thread pool minimum size `Node-Zero' is updated to `6242'. \textit{(distractor)}} 
    \textcolor{black}{...(context)...}
    \textcolor{cyan!60!black}{[System Log Seq 005] The thread pool minimum size `Echo-Base' is updated to `5666'. \textit{(indirect)}} 
    \textcolor{red!75!black}{[System Log Seq 000] The thread pool minimum size `Node-Zero' is initially set to `4259'. \textit{(distractor)}} 
    \textcolor{black}{...(context)...}
    \textcolor{cyan!60!black}{[System Log Seq 008] The thread pool minimum size `Echo-Base' is updated to `9115'. \textit{(indirect)}} 
    \textcolor{red!75!black}{[System Log Seq 001] The thread pool minimum size `Node-Zero' is updated to `5387'. \textit{(distractor)}} 
    \textcolor{black}{...(context)...}
    \textcolor{cyan!60!black}{[System Log Seq 006] The thread pool minimum size `Echo-Base' is updated to `1107'. \textit{(indirect)}} 
    \textcolor{black}{...(context)...}
    \textcolor{cyan!60!black}{[System Log Seq 004] ... \textit{(indirect)}} 
    \textcolor{red!75!black}{[System Log Seq 004] The thread pool minimum size `Node-Zero' is updated to `8654'. \textit{(distractor)}} 
    \textcolor{black}{...(context)...}
    \textcolor{green!50!black}{System architecture documentation confirms that the internal alias `Echo-Base' represents the `log\_level', and the `thread pool minimum size' structurally signifies the configuration variable. \textit{(direct)}} 
    \textcolor{black}{...(context)...}
    \textcolor{red!75!black}{[System Log Seq 002] ... \textit{(distractor)}}
    \textcolor{cyan!60!black}{\textbf{[System Log Seq 009] The thread pool minimum size `Echo-Base' is updated to `1434'. \textit{(indirect)}}} 
    \textcolor{black}{...(context)...}
    \textcolor{cyan!60!black}{[System Log Seq 007] ... \textit{(indirect)}}
    \textcolor{black}{...(context)...}} \\
    \bottomrule
    \end{tabular}

    \caption{Two cases of the Global Reasoning Task. Each case is presented in a boxed single-column format, where the direct fact is highlighted in \textcolor{green!50!black}{green}, distractors in \textcolor{red!75!black}{red}, and indirect facts in \textcolor{cyan!60!black}{cyan}.}
    \label{tab:preliminary_cases}
\end{table*}

\onecolumn

\begin{table*}[htbp]
    \centering
    \setlength{\tabcolsep}{6pt}
    \renewcommand{\arraystretch}{1.25}
    \begin{tabular}{p{0.95\linewidth}}
    \toprule
    \multicolumn{1}{c}{\textbf{2WikiMultiHopQA Case: Compositional Bridge Reasoning}} \\
    \hline
    \textbf{Question}: Who is the father of the director of film \textit{The Seven Madmen}? \\
    \midrule
    \textbf{Answer}: Leopoldo Torres Ríos \\
    \midrule
    \textbf{Document}: {\footnotesize
    \textcolor{black}{...(context)...}
    \textcolor{red!75!black}{\textit{Mexican Spitfire Out West} is a 1940 American comedy film directed by Leslie Goodwins and written by Charles E. Roberts and Jack Townley. \textit{(distractor)}} 
    \textcolor{black}{...(context)...}
    \textcolor{cyan!60!black}{Leopoldo Torre Nilsson, also known as Leo Towers and Babsy, was an Argentine film director, producer, and screenwriter. Born as Leopoldo Torres Nilsson, he was the son of Argentine pioneer film director Leopoldo Torres Ríos. \textit{(indirect)}} 
    \textcolor{black}{...(context)...}
    \textcolor{red!75!black}{Anthony Chinn was a British supporting actor who appeared in over 50 films and television series. He was the child of Chinese and Brazilian parents. \textit{(distractor)}} 
    \textcolor{black}{...(context)...}
    \textcolor{green!50!black}{\textit{The Seven Madmen}, also known as \textit{The Revolution of the Seven Madmen}, is a 1973 Argentine drama film directed by Leopoldo Torre Nilsson. \textit{(direct)}} 
    \textcolor{black}{...(context)...}} \\
    \bottomrule
    \end{tabular}
    \caption{A representative 2WikiMultiHopQA case exhibiting the non-linear pattern. The direct fact is highlighted in \textcolor{green!50!black}{green}, distractors in \textcolor{red!75!black}{red}, and indirect facts in \textcolor{cyan!60!black}{cyan}.}
    
    \label{tab:2wiki_non_linear_case}
\end{table*}

% \onecolumn
\begin{algorithm}[H]
\caption{ MemReread Working Mechanism.}
\label{alg:mem_reread}
\begin{algorithmic}[1]
    \State \textbf{Require:} Backbone Model $LLM$, reading template $T_R$, answering template $T_A$, decomposing template $T_D$ and integrating templte $T_I$ 
    \Function{MemorizeWhileReading}{$q$, $C$} \Comment{\textcolor{gray}{$q$ is the question. $C$ is the list of context chunks $c_i, i =0, 1, ..., T-1$.}}

    \State $m$ = NO\_MEMORY
    \For{$c$ in $C$}
    \State $m \leftarrow LLM(T_R(q,m,c))$ 
    
    \EndFor

    \State \Return $m$

    \EndFunction
\\
    \Function{Answer}{$q$, $m$} \Comment{\textcolor{gray}{$q$ is the question. $m$ is the final memory after reading all chunks.}}

    \State $a \leftarrow LLM(T_A(Q, m, c))$ 
    \State \Return $a$
    \EndFunction

\\

\Function{\textbf{MemReread}}{$Q$, $C$, $p$} \Comment{\textcolor{gray}{$Q$ is the question. $C$ is the list of $c_i, i =0, 1, ..., T-1$. $p$ is the rereading passes limit.}}
    % \State \textbf{Output:} $NULL$
    \State $M \leftarrow$ \textsc{MemorizeWhileReading}($Q$, $C$)
    \State $qa \leftarrow [\ ]$ \Comment{\textcolor{gray}{$qa$ is a list of historical subquestion-answers.}}
    \For{ $i = 1$ to $p$}
    \State $d \leftarrow LLM(T_D(Q, M, qa))$
    \If{ \textbf{not} \textsc{HasQuestion}($d$)}\Comment{\textcolor{gray}{Rule-based question matching.}}
        \State \textbf{break}
    \EndIf

    \State $q \leftarrow$ \textsc{ParseQuestion}($d$)\Comment{\textcolor{gray}{Rule-based question parsing.}}
    
    \State $m \leftarrow$ \textsc{MemorizeWhileReading}($q$, $C$)
    \State $a \leftarrow$ \textsc{Answer}($q$, $m$)

    \State $M \leftarrow LLM(T_I(Q, M, q, a))$
    \State $qa \leftarrow qa + [(q, a)]$
    \EndFor
    
    \State $A \leftarrow$ \textsc{Answer}($Q$, $M$)
    \State \Return $A$

\EndFunction
\end{algorithmic}
\end{algorithm}

\twocolumn

\section{Implementation Details}
\label{app:design_logic}

\subsection{Prompt Template}
\label{app:prompt_template}

We provide the complete set of four prompt templates used in MemReread. The \textit{Reading} and \textit{Answering} templates are adopted verbatim from MemAgent without modification, as illustrated in Figure~\ref{fig:ra_template}. For the \textit{Decomposing} template, we avoid tool calling to ensure broader compatibility; instead, we instruct the model to enclose sub-questions within \texttt{<query></query>} tags. We then employ rule-based parsing on these tags to determine whether a sub-question has been generated, which dictates whether rereading is required. For the \textit{Integrating} template, we update the memory solely with the sub-question and the sub-answer (excluding the sub-memory). Following integration, we append the sub-question with its answer to the question-answer history to mitigate redundant sub-question decomposition. Additional details are provided in Figure~\ref{fig:di_template}.

\begin{figure*}[htbp]
\begin{AcademicBox}
% [Reading/Answering Template]
\footnotesize
\textbf{\textit{Reading:}} \\
        You are presented with a problem, a section of an article that may contain the answer to the problem, and a previous memory. Please read the provided section carefully and update the memory with the new information that helps to answer the problem. Be sure to retain all relevant details from the previous memory while adding any new, useful information.

        \texttt{<problem>} \{question\} \texttt{</problem>} \textbackslash n
        \texttt{<memory>} \{memory\} \texttt{</memory>} \textbackslash n
        \texttt{<section>} \{chunk\} \texttt{</section>}

        Updated memory: \\

\vspace{-5pt} \hrule \vspace{4pt}

\textbf{\textit{Answering:}} \\
        You are presented with a problem and a previous memory. Please answer the problem based on the previous memory and put the answer in \textbackslash boxed\{\}. \\
        \texttt{<problem>} \{question\} \texttt{</problem>} \textbackslash n
        \texttt{<memory>} \{memory\} \texttt{</memory>}\\
        Your answer: 
\end{AcademicBox}

\caption{Reading and Answering Template}

\label{fig:ra_template}
\end{figure*}

\begin{figure*}[htbp]
\begin{AcademicBox}
% [Decomposing/Integrating Template]
\footnotesize
\textbf{\textit{Decomposing:}} \\
You are a Question Analysis and Query Generation Agent. 
You will receive a QUESTION, a MEMORY containing known facts and evidence, and a QUERY\_HISTORY containing previously submitted exploratory queries with their results. Your task is to analyze whether the information in the MEMORY is sufficient to fully answer the QUESTION. If it is insufficient, you must generate a single targeted exploratory query to fill the most critical gap.

Critical Rules \& Analysis Steps:\\
1. Focus on QUESTION: Evaluate only whether MEMORY can answer the current QUESTION. Do not attempt to answer or re-query any other questions.\\
2. Compare needs with MEMORY: Break the QUESTION down into specific information needs and compare them with what the MEMORY already contains. Identify if any crucial facts are still missing, uncertain, or incomplete.\\
3. Review QUERY\_HISTORY: Avoid repeating any query already asked and recorded in QUERY\_HISTORY.\\
4. You MUST NOT answer the QUESTION directly. Your output must strictly follow the decision logic below.\\
Decision Logic \& Actions:\\
- IF SUFFICIENT: If the MEMORY already contains all necessary information to fully address the QUESTION, you must stop immediately. Do not generate any query and do not generate any further text.\\
- IF INSUFFICIENT: If the MEMORY lacks necessary information for the QUESTION, you must leverage the known information in the MEMORY to identify and submit the single highest-priority exploratory query that must be resolved first to push the QUESTION forward.\\
Guidelines for the Priority Query (Only if MEMORY is insufficient for the QUESTION):
1. Highest Priority: Among all possible information gaps for the QUESTION, select only the one query whose resolution is most critical and foundational for answering the QUESTION. \\
2. No Repetition: The query must NOT duplicate any query already present in the QUERY\_HISTORY. Review the history carefully and ensure your proposed query is novel and represents genuine progress.\\
3. Independence: The query must be answerable in isolation without requiring answers to other queries.\\
4. Sufficiency: Resolving this query, combined with the existing MEMORY, must meaningfully progress toward fully resolving the QUESTION.\\
5. Self-Contained Expression: The query must be fully self-contained and free of any references to the original QUESTION, MEMORY, or external context. Never use pronouns, option labels, or context-dependent phrases like "the entity mentioned above", "option A", or "this event". Instead, explicitly state all relevant entities, values, and content.\\
6. Output Format: Submit exclusively this single highest-priority query by wrapping it in <query> tags. Output exactly one query; never submit multiple queries.\\
7. Confirmed Information Exclusion: Do NOT generate a query targeting any information enclosed within <confirmed>...</confirmed> tags in the MEMORY. These tags denote facts, evidence, or confirmed absences that have already been explicitly verified and resolved. Only generate queries for information whose existence, value, or status remains uncertain, ambiguous, or entirely unaddressed in the MEMORY. This prevents redundant exploration of settled information and directs attention to genuinely unresolved gaps.\\
        QUESTION that you need to focus on: \\
        \texttt{<question>} \{question\} \texttt{</question>} \textbackslash n \texttt{<memory>} \{memory\} \texttt{</memory>} \\
        \texttt{<query\_history>} \{qa\_history\} \texttt{</query\_history>} \\

\vspace{-5pt} \hrule \vspace{4pt}
\textbf{\textit{Integrating:}} \\
You are a memory integration assistant. You will receive three inputs: a QUESTION, a current MEMORY snippet, and a reference subquestion-answer pair obtained from the latest question progression step. Your task is to integrate the information from the reference into the MEMORY so that the updated MEMORY progresses towards answering the original QUESTION. Follow these rules strictly:\\
1. Do not answer the original QUESTION directly; your sole output is the integrated MEMORY.\\
2. When information in the reference conflicts with existing MEMORY content, prioritize the reference information as it represents fresh research.\\
3. Eliminate redundant information; if a fact already exists in MEMORY, do not add it again.\\
4. Filter out any information irrelevant to the original QUESTION; retain only content that contributes to answering it.\\
5. Express all integrated information in fluent, coherent natural language. Do not copy the reference verbatim; instead, extract key facts and synthesize them into descriptive prose.\\
6. Cross-Source Evidence Tagging: Wrap a statement in <confirmed>...</confirmed> tags if the exact same factual claim (regarding existence, occurrence, or verified absence of evidence) appears in BOTH the current <memory> input AND the <subanswer> section of the reference. This cross-source consistency indicates higher reliability. Do NOT tag information that appears in only one of these two sources, or that is speculative, uncertain, or inferred.\\
7. Preserve Existing Confirmed Tags: If the current <memory> already contains information enclosed in <confirmed>...</confirmed> tags, and this information is not contradicted by the reference, retain these tagged segments verbatim in the updated MEMORY. Integrate them naturally into the surrounding prose without removing the tags.\\
Your final output should be a concise, well-structured MEMORY that consolidates all verified, relevant information needed to resolve the original QUESTION.\\
    \texttt{<question>} \{question\} \texttt{</question>} \textbackslash n \texttt{<memory>} \{memory\} \texttt{</memory>} \textbackslash n
    \texttt{<reference>} \textbackslash n
    \texttt{<subquestion>} \{subquestion\} \texttt{</subquestion>} \textbackslash n
    \texttt{<subanswer>} \{subanswer\} \texttt{</subanswer>} \textbackslash n
    \texttt{</reference>} \\
    Updated memory: 
% \caption{Decomposing and Integrating Template}
\end{AcademicBox}

\caption{Decomposing and Integrating Template}
\label{fig:di_template}

\end{figure*}
%     \end{tabular}
% \end{table}

\subsection{Algorithm}

By unifying each iteration of sub-question generation, question-guided rereading, and memory-based sub-answer derivation into a single function call, our approach extends the ReAct~\cite{yao2023react} paradigm to memory management in long-context reasoning scenarios. The workflow is detailed in Algorithm~\ref{alg:mem_reread}.

\subsection{Training Details}
\label{app:training}
\subsubsection{Step-Level Advantage}
\label{app:training_adv}

We adopt the state reward formulation from ReMemR1 to provide denser process supervision. Notably, given the iterative rereading procedure, we compute the reward at each reading pass:

\begin{equation}
\begin{aligned}
R_{\mathrm{state},p,t}^{(g)}
&=
\max\nolimits_{y\in Y}
\operatorname{recall}
\bigl(m_{p,t}^{(g)},y\bigr)
\\[-0.2em]
&\quad -
\max\nolimits_{y\in Y}
\operatorname{recall}
\bigl(m_{p,t-1}^{(g)},y\bigr).
\end{aligned}
\label{eq:ab_rewards}
\end{equation}
where $Y$ denotes the set of golden answers, and $p$ denotes the number of completed reading passes.

Finally, the step-level advantage is computed as:

\begin{equation}
    \begin{aligned}
        \hat{A}^{(g)}_{\text{state}, p, t} &= R^{(g)}_{\text{state}, p, t} - \frac{1}{G} \sum_{k =1}^{G}R_{\text{state},p,t}^{(k)}
    \end{aligned}
\label{eq:state_rewards}
\end{equation}
Notably, given the different number of reading passes across trajectories, the process-level advantage for any reading steps extending beyond the common trajectory length is explicitly set to zero. For these excess steps, the advantage is determined exclusively by the outcome-level advantage.

\subsubsection{Training Objective}
\label{app:training_obj}
The full expression of our training objective can be written as:

\begin{equation}
\small
\begin{aligned}
&\operatorname*{arg\,max}_{\theta}
J_{\text{ReA-GRPO}}(\theta) =
\\[-0.2em]
&\quad 
\operatorname*{arg\,max}_{\theta}
\mathbb{E}_{(Q,Y),\{\tau^{(g)}\}_{g=1}^{G} \sim \pi_{\theta_{\text{old}}}} \Bigg[
\\
&\qquad
\frac{1}{G(T+1)}
\sum\nolimits_{g=1}^{G}
\sum\nolimits_{t=1}^{T+1}
\sum\nolimits_{p=0}^{p^{(g)}}
\frac{1}{\lvert s_{p,t}^{(g)}\rvert}
\sum\nolimits_{i=1}^{\lvert s_{p,t}^{(g)}\rvert}
\\
&\qquad
\min\Bigg(
  \rho_{p,t,i}^{(g)}
  \hat{A}_{p,t}^{(g)},
  \operatorname{clip}\!\left(
    \rho_{p,t,i}^{(g)},
    1-\epsilon,
    1+\epsilon
  \right)
  \hat{A}_{p,t}^{(g)}
\Bigg)
\\
&\qquad
-
\beta
D_{\mathrm{KL}}\!\left[
  \pi_{\theta}
  \,\middle\|\,
  \pi_{\mathrm{ref}}
\right]
\Bigg].
\end{aligned}
\label{eq:rea_grpo}
\end{equation}

where $\rho_{p, t,i}^{(g)}$ is the importance sampling ratio:

\begin{equation}
\rho_{p, t,i}^{(g)} = \frac{\pi_{\theta}\left(s_{p, t,i}^{(g)} \mid s_{p, t,<i}^{(g)}, s_{p, <t}^{(g)}, s_{<p}^{(g)}, Q, c_{t-1}\right)}{\pi_{\theta_{\text{old}}}\left(s_{p, t,i}^{(g)} \mid s_{p, t,<i}^{(g)}, s_{p, <t}^{(g)}, s_{<p}^{(g)}, Q, c_{t-1}\right)}.
\end{equation}

\subsubsection{Training Configurations}
\label{app:training_hyperparameters}
Our training pipeline is built upon the \texttt{Verl}\footnote{https://github.com/verl-project/verl} framework, employing \texttt{vLLM}\footnote{https://github.com/vllm-project/vllm} as the rollout engine. The complete set of training hyperparameters is summarized in Table~\ref{tab:training_hyperparameters}. All our models were trained on 8 $\times$ NVIDIA A800 (80G) GPUs. Specifically, our 1.7B model converges after 20 hours, while our 4B model converges after 40 hours.

\begin{table}[htbp]
% \begin{wraptable}{r}{0.4\textwidth}
    \centering
    \centering
    % \resizebox{0.4\textwidth}{!}{
    \begin{tabular}{@{}ll@{}}
      \toprule
      \textbf{Hyperparameter} & \textbf{Value} \\
      \midrule
      Training Batch Size         & 64 \\
      Micro Training Batch Size    & 8 \\
      Total Convergence Steps    & 80 \textasciitilde 120 \\
      Learning Rate              & 1e-6 \\
      Warmup Steps             & 20 \\
      Rollout Temperature     & 1.0 \\
      Max Chunk Length        & 5000 \\
      Max Chunk Number $T$        & 8 \\
      Max Rereading Pass $p_c$      & 2 \\
      Max Response Length     & 1024 \\
      Outcome Reward Weight $\alpha$ & 0.95 \\
      KL Coefficient $\beta$         & 0.001 \\
      Clip Ratio $\epsilon$         & 0.02 \\
      Group Size $G$             & 4 \\
      
      \bottomrule
    \end{tabular}
    % }
    \caption{Primary hyperparameters used in Training}
    \label{tab:training_hyperparameters}
% \end{wraptable}
\end{table}

\subsection{Evaluation Details}
\label{app:evaluation}
All evaluations(except API-based ones) were run on 4 $\times$ NVIDIA A800 (80G) GPUs. Given the extraordinary computational cost of long context evaluation in the main experiments, we subsample 128 samples per context length for 2WikiMultihopQA, following~\cite{yu2026memagent}. For additional benchmarks in Appendix~\ref{app:more_benchmarks}, we evaluate their QA/reasoning tasks. Specifically, we test the full LongBench-v2 dataset with a 1M maximum context length (truncating any excess). For RULER-QA, we subsample 64 samples per length across contexts ranging from 8K to 1M.

% \clearpage
\section{Additional Results}
\label{app:supplementary_exp}

\subsection{In-Distribution Performance}

We report the results on HotPotQA~\cite{yang2018hotpotqa} compared to advanced RL-augmented memory agents in Table~\ref{tab:main_result_additional}.
\begin{table*}[t]
\centering
\vspace{-0.5em}
% \resizebox{\textwidth}{!}{
    \small
    \begin{tabular}{c l | c c c c c c c c | c}
% \begin{tabularx}{\textwidth}{c | C C C C C | C C C C C }
        \toprule
        \multirow{2.5}{*}{\textbf{Scale}} & \multirow{2.5}{*}{\textbf{Framework}} & \multicolumn{8}{c|}{\textbf{\emph{Context Length}}} & \multirow{2.5}{*}{\textbf{Avg.}}  \\
        \cmidrule(lr){3-10}
        & & 8K & 16K & 32K & 64K & 128K & 256K & 512K & 1M  &\\
        \arrayrulecolor{black}\midrule
        % \rowcolor{orange!6} \multicolumn{11}{c}{\textit{\textbf{HotpotQA (In-Distribution)}}} \\
        \arrayrulecolor{black!20}\midrule
        
        \multirow{4}{*}{1.7B} &MemAgent & \textbf{44.5} & 39.1 & 29.7 & \textbf{23.4} & 25.0 & 20.3 & 21.1 & \textbf{19.5} & 27.8 \\
        & ReMemR1 & 27.3 & 28.1 & 31.2 & 19.5 & \textbf{25.8} & \textbf{25.8} & 21.9 & 17.2  & 24.6 \\
        & \textbf{MemReread(Ours)} & 43.0 & \textbf{43.0} & \textbf{32.0} & \textbf{23.4} & 22.7 & 22.7 & \textbf{23.4} & 17.2 & \textbf{28.4} \\
        
        \arrayrulecolor{gray}\midrule
        \multirow{4}{*}{4B}
        & MemAgent & 53.9 & 55.5 & 46.9 & \textbf{50.8} & 43.8 & 43.0 & 43.8 & 42.2 & 47.5 \\
        & ReMemR1 & 58.6 & 54.7 & \textbf{57.0} & 50.0 & \textbf{51.6} & 45.3 & 50.0 & 52.3  & 52.4 \\
        & \textbf{MemReread(Ours)} & \textbf{59.4} & \textbf{57.8} & 51.6 & 46.9 & 49.2 & \textbf{53.1} & \textbf{51.6} & \textbf{54.7} & \textbf{53.0} \\
        \arrayrulecolor{black}\toprule
% \end{tabularx}
    \end{tabular}
\caption{Results on HotpotQA~\cite{yang2018hotpotqa}. Values are accuracy(\%), rounded to 1 decimal. \textbf{Bold} denotes the best performances.}
% }
\label{tab:main_result_additional}
\end{table*}

\subsection{Computation Overhead}
\label{app:subsectoin:overhead}

We report the detailed results of Figure~\ref{fig:time_cost}
in Table~\ref{tab:comprehensive_performance_overhead}.

\begin{table*}[htbp]
\centering
\vspace{-0.5em}
\small
\resizebox{\textwidth}{!}{%
\begin{tabular}{l c l | c c c c c c c c}
    \toprule
    \multirow{2.5}{*}{\textbf{Framework}}
    & \multirow{2.5}{*}{\textbf{Max New Tokens}}
    & \multirow{2.5}{*}{\textbf{Metric}}
    & \multicolumn{8}{c}{\textbf{\emph{Context Length}}} \\
    \cmidrule(lr){4-11}
    & & & 8K & 16K & 32K & 64K & 128K & 256K & 512K & 1M \\
    \midrule

    \multirow{3}{*}{MemAgent}
    & \multirow{3}{*}{1024}
    & Accuracy (\%) & 68.0 & 51.6 & 43.8 & 39.1 & 39.8 & 39.1 & 35.2 & 39.8 \\
    & & Time (s) & 0.344 & 0.688 & 1.375 & 2.750 & 5.549 & 11.102 & 22.037 & 44.021 \\
    & & Memory (MB) & 0.009 & 0.010 & 0.013 & 0.009 & 0.013 & 0.010 & 0.015 & 0.011 \\
    \midrule

    \multirow{3}{*}{ReMemR1}
    & \multirow{3}{*}{2048}
    & Accuracy (\%) & 64.1 & 54.7 & 44.5 & 37.5 & 41.4 & 49.2 & 37.5 & 41.4 \\
    & & Time (s) & 0.406 & 0.813 & 1.625 & 3.250 & 6.511 & 13.008 & 26.043 & 52.101 \\
    & & Memory (MB) & 0.010 & 0.022 & 0.032 & 0.050 & 0.100 & 0.165 & 0.250 & 0.392 \\
    \midrule

    \multirow{3}{*}{\textbf{MemReread (Ours)}}
    & \multirow{3}{*}{1024}
    & Accuracy (\%) & 70.3 & 71.1 & 59.4 & 64.1 & 54.7 & 55.5 & 46.9 & 45.3 \\
    & & Time (s) & 1.317 & 2.549 & 5.046 & 10.127 & 20.809 & 40.855 & 79.113 & 157.595 \\
    & & Memory (MB) & 0.010 & 0.010 & 0.010 & 0.009 & 0.010 & 0.016 & 0.013 & 0.015 \\
    \midrule

    \multirow{3}{*}{\textbf{MemReread-Flash (Ours)}}
    & \multirow{3}{*}{1024}
    & Accuracy (\%) & 69.5 & 62.5 & 58.6 & 53.9 & 44.5 & 51.6 & 42.2 & 42.7 \\
    & & Time (s) & 0.957 & 1.331 & 2.020 & 3.736 & 7.373 & 15.038 & 30.213 & 58.622 \\
    & & Memory (MB) & 0.009 & 0.009 & 0.008 & 0.009 & 0.008 & 0.014 & 0.012 & 0.013 \\
    \bottomrule
\end{tabular}%
}
\caption{Detailed results of Figure~\ref{fig:time_cost}.}
\label{tab:comprehensive_performance_overhead}
\end{table*}

\begin{table}
    \centering
    \resizebox{0.45\textwidth}{!}{
        \begin{tabular}{l | c}
            \toprule
            \textbf{Framework(4B scale)} & \textbf{P-Value} \\
            \arrayrulecolor{black}\midrule
            \textbf{MemReread (Ours)} V.S. MemAgent & 1.868e-14 \\
            \textbf{MemReread (Ours)} V.S. ReMemR1 & 2.474e-13 \\
            \bottomrule
        \end{tabular}
    } % 注意这里括号的闭合
    \caption{Statistical significance calculation on \texttt{2WikiMultihopQA} with paired t-test.}
    \label{tab:statistical_significance}
\end{table}

\subsection{Comprison with GRPO}
\begin{figure*}[htbp]
    \centering
    \includegraphics[width=0.8\linewidth]{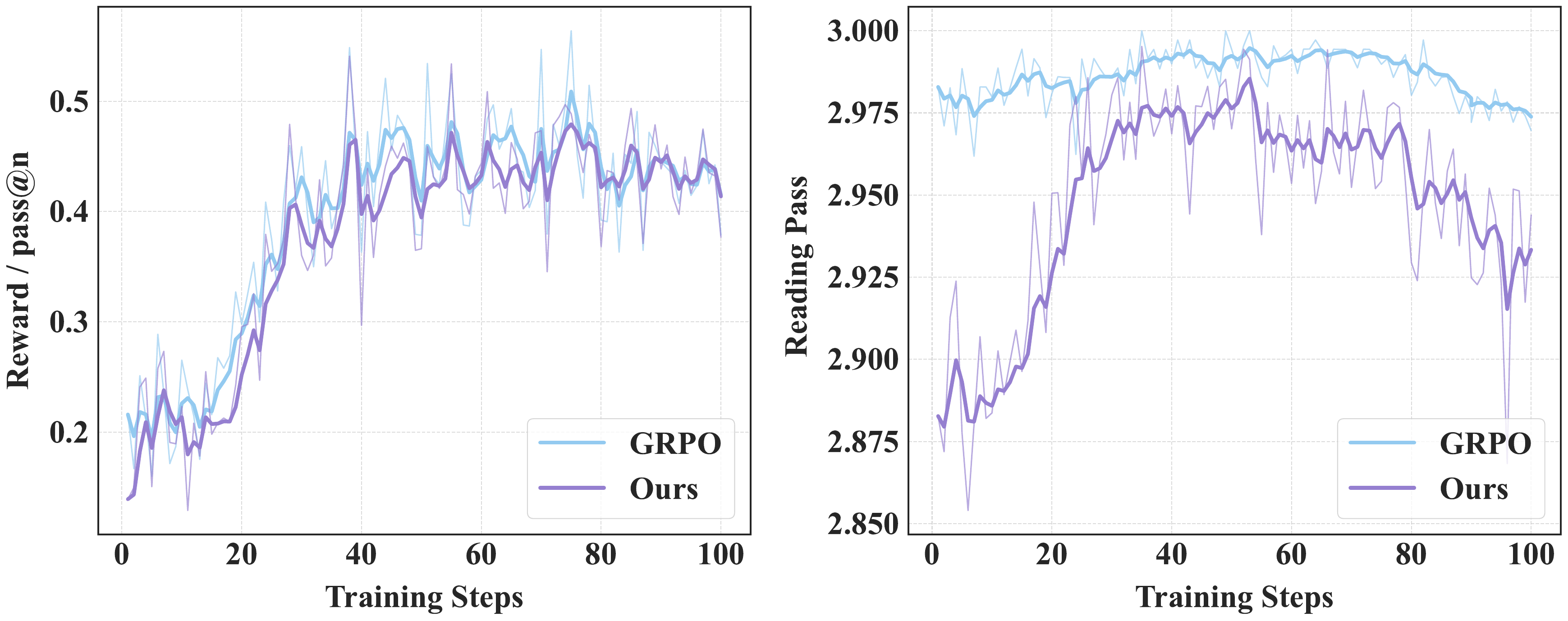}
    \caption{Comparison of GRPO and ReA-GRPO(Ours)}
    \label{fig:training_curve}
\end{figure*}

As shown in Figure~\ref{fig:training_curve}, ReA-GRPO matches GRPO's validation performance with a modest reduction in average reading passes. This slight efficiency gain stems from our advanced design, which triggers extra readings exclusively for challenging samples. This selective mechanism aligns with the multi-hop nature, in which targeted revisits are essential for correctness, enabling ReA-GRPO to maintain accuracy while reducing unnecessary rereading steps. 

The numerical values of each context length on 2WikimultihopQA in Figure~\ref{fig:rl_comparison_benchmark} are reported in Table~\ref{tab:rl_comparison_number}.

\begin{table*}[htbp]
\centering
\vspace{-0.5em}
\small
\resizebox{\textwidth}{!}{%
\begin{tabular}{l l | c c c c c c c c}
    \toprule
    \multirow{2.5}{*}{\textbf{Method}} & \multirow{2.5}{*}{\textbf{Metric}} & \multicolumn{8}{c}{\textbf{\emph{Context Length}}} \\
    \cmidrule(lr){3-10}
    & & 8K & 16K & 32K & 64K & 128K & 256K & 512K & 1M \\
    \midrule
    \multirow{2}{*}{Vanilla} 
    & Accuracy (\%) & 60.9 & 57.0 & 50.8 & 46.1 & 36.7 & 39.8 & 32.0 & 33.6  \\
    & Average Rereading (p) & 2.85 & 2.82 & 2.95 & 2.85 & 2.91 & 2.97 & 3.00 & 3.00 \\
    \midrule
    \multirow{2}{*}{~~~+ GRPO} 
    & Accuracy (\%) & 62.5 & 61.7 & 64.1 & 46.9 & 39.8 & 46.1 & 34.4 & 43.0 \\
    & Average Rereading (p) & 2.93 & 2.89 & 2.94 & 2.96 & 2.93 & 2.96 & 2.95 & 2.91 \\
    \midrule
    \multirow{2}{*}{~~~+ \textbf{ReA-GRPO (Ours)}} 
    & Accuracy (\%) & 70.3 & 71.1 & 59.4 & 64.1 & 54.7 & 55.5 & 46.9 & 45.3 \\
    & Average Rereading (p) & 2.81 & 2.90 & 2.84 & 2.87 & 2.84 & 2.87 & 2.89 & 2.85 \\
    \bottomrule
\end{tabular}%
}
\caption{Comparison of MemReread-4B with different RL frameworks on 2WikiMultiHopQA, where we set $p_c=3$. Vanilla denotes no RL training.}
\label{tab:rl_comparison_number}
\end{table*}

\begin{table*}[t]
\centering
\vspace{-0.5em}
\small
\resizebox{\textwidth}{!}{
\begin{tabular}{l l | c c c}
    \toprule
    \multirow{2.5}{*}{\textbf{Framework}} & \multirow{2.5}{*}{\textbf{Metric}} & \multicolumn{3}{c}{\textbf{\emph{Benchmark}}} \\
    \cmidrule(lr){3-5}
    & & RULER-QA & LongBench-E-QA & LongBenchv2 \\
    \midrule
    \multirow{2}{*}{Vanilla} 
    & Score (\%) & 54.8 & 47.2 & 27.1 \\
    & Average Rereading (p) & 2.03 & 2.50 & 2.61 \\
    \midrule
    \multirow{2}{*}{~~~+ GRPO} 
    & Score (\%) & 58.8 & 50.0 & 30.2 \\
    & Average Rereading (p) & 2.42 & 2.86 & 2.43 \\
    \midrule
    \multirow{2}{*}{~~~+ \textbf{ReA-GRPO(Ours)}} 
    & Score (\%) & 64.0 & 50.9 & 32.1 \\
    & Average Rereading (p) & 0.17 & 2.91 & 2.24 \\
    \bottomrule
\end{tabular}
}
\caption{Detailed results of Figure~\ref{fig:rl_comparison_benchmark}. Comparison of MemReread-4B with different RL frameworks on more benchmarks, where we set $p_c=3$. Vanilla denotes no RL training.}
\label{tab:benchmark_performance}
\end{table*}

\subsection{Comparison with Additional Baselines}
\label{app:compare_base}

Previous works~\cite{yu2026memagent,shi2026look} have already established the advantage of memory agents over LLMs in ultra-long-context scenarios. Here we briefly report our performance gains on the 4B scale. As shown in Table~\ref{tab:compare_qwen3}, our method consistently outperforms LLM.

Furthermore, we compare our approach with CDT~\cite{tang2026revisiting}, a representative post-training strategy to enhance long context performance, reproduced using LongAlpaca-12K~\cite{chen2024longlora} (capped at 40K to match our setup). As shown in Table~\ref{tab:compare_qwen3}, despite gains on 2WikiMultiHopQA at 8K and 16K, CDT suffers significant degradation beyond 32K. We attribute this to SFT compromising the model's inherent reasoning capabilities~\cite{lobo2025impact}. Specifically, SFT biases the model toward surface-level pattern matching, which suffices for shorter contexts but fails to sustain multi-hop tracking in long contexts.

We also benchmark our method against InfMem, a chunk-retrieval approach. Given that only 4B-parameter checkpoints are publicly available for InfMem, we conduct evaluations at 4B scale on out-of-distribution (OOD) datasets to ensure a rigorous comparison. As shown in Table~\ref{tab:compare_infmem}, MemReread demonstrates better overall performance. Notably, during streaming inference, InfMem's context window expands beyond 40K tokens, incurring prohibitive computational and memory overhead. In contrast, our framework caps the active context length at less than 8K tokens at all times.

\begin{table*}[htbp]
\centering
\vspace{-0.5em}
% \resizebox{\textwidth}{!}{
    \small
    \begin{tabular}{c l | c c c c c c c c | c}
% \begin{tabularx}{\textwidth}{c | C C C C C | C C C C C }
        \toprule
        \multirow{2.5}{*}{\textbf{Scale}} & \multirow{2.5}{*}{\textbf{Framwork}} & \multicolumn{8}{c|}{\textbf{\emph{Context Length}}} & \multirow{2.5}{*}{\textbf{Avg.}}  \\
        \cmidrule(lr){3-10}
        & & 8K & 16K & 32K & 64K & 128K & 256K & 512K & 1M  &\\
        \arrayrulecolor{black}\midrule

         \multirow{2.5}{*}{4B}& Qwen3 & 67.2 & 60.9 & 52.3 & 43.8 & 46.9 & - & - & -  & - \\
         & ~~~+ {CDT}~\cite{tang2026revisiting} & \textbf{73.4} & 61.7 & 44.5 & 39.8 & 37.5 & - & - & -  & - \\
        &  ~~~+ \textbf{MemReread} & 70.3 & \textbf{71.1} & \textbf{59.4} & \textbf{64.1} & \textbf{54.7} & \textbf{55.5} & \textbf{46.9} & \textbf{45.3} & \textbf{58.4} \\
        \bottomrule
% \end{tabularx}
    \end{tabular}
% }
\caption{Comparison with LLM on 2WikiMultiHopQA~\cite{ho2020constructing}.}
\label{tab:compare_qwen3}
\end{table*}

\begin{table*}[htbp]
\centering
\vspace{-0.5em}
% \resizebox{\textwidth}{!}{
    \small
    \begin{tabular}{c l | c c c c c c c c | c}
% \begin{tabularx}{\textwidth}{c | C C C C C | C C C C C }
        \toprule
        \multirow{2.5}{*}{\textbf{Scale}} & \multirow{2.5}{*}{\textbf{Framwork}} & \multicolumn{8}{c|}{\textbf{\emph{Context Length}}} & \multirow{2.5}{*}{\textbf{Avg.}}  \\
        \cmidrule(lr){3-10}
        & & 8K & 16K & 32K & 64K & 128K & 256K & 512K & 1M  &\\
        \arrayrulecolor{black}\midrule
        \multirow{2.5}{*}{4B}& InfMem~\cite{wang2026infmem} & 53.1 & 62.5 & 55.5 & 53.1 & \textbf{55.5} & \textbf{68.0} & \textbf{53.1} & \textbf{49.2} & 56.3 \\
        & \textbf{MemReread} & \textbf{70.3} & \textbf{71.1} & \textbf{59.4} & \textbf{64.1} & 54.7 & 55.5 & 46.9 & 45.3 & \textbf{58.4} \\
        \bottomrule
% \end{tabularx}
    \end{tabular}
% }
\caption{Comparison with InfMem on 2WikiMultiHopQA~\cite{ho2020constructing}.}
\label{tab:compare_infmem}
\end{table*}

\subsection{Experiment Statistical Significance}

\label{section:exp_stat_significance}

We perform paired-sample t-tests using sample-level predictions from the 2WikiMultihopQA test dataset. With correct and incorrect predictions represented as 1 and 0, respectively, MemReread significantly outperforms both ReMemR1 and MemAgent ($p < 0.05$), as shown in Table~\ref{tab:statistical_significance}. These results confirm that the performance improvements of MemReread are statistically significant.

\subsection{Comparison on Additional Benchmarks}
\label{app:more_benchmarks}

We further evaluate our method on widely adopted long-context benchmarks, as shown in Tables~\ref{tab:ruler_results}, \ref{tab:longbench_all}, \ref{tab:longbench_v2}. Our approach consistently outperforms all baselines in all benchmarks. Notably, the performance advantage is most pronounced in ultra-long-context subsets, such as RULER-QA ($>$256K) and LongBench-v2-1M(Long). However, the performance margin narrows on shorter tasks like LongBench-QA and LongBench-E-QA, where maximum context lengths are capped at approximately 20K tokens. We attribute this less obvious gap to the reduced chunk granularity in shorter sequences: with at most four context chunks, the probability of losing critical cross-chunk dependencies is inherently low. Consequently, performance in these scenarios is predominantly governed by the backbone model's intrinsic reasoning capability rather than context management capability. Given that our framework and all baselines share the identical backbone architecture, a performance alignment on short-context tasks is expected. This observation further underscores that our method's core advantage lies in preserving long-range dependencies and mitigating information fragmentation across extended context windows.

\begin{table*}[htbp]
\centering
\vspace{-0.5em}
\small
\begin{tabular}{l | c c c c c c c c | c}
    \toprule
    \multirow{2.5}{*}{\textbf{Framework}} & \multicolumn{8}{c|}{\textbf{\emph{Context Length}}} & \multirow{2.5}{*}{\textbf{Avg.}} \\
    \cmidrule(lr){2-9}
    & 8K & 16K & 32K & 64K & 128K & 256K & 512K & 1M & \\
    \arrayrulecolor{black}\midrule
    MemAgent & 67.2 & 69.5 & 53.1 & 51.6 & 51.6 & 50.8 & 45.3 & 57.0 & 55.8 \\
    ReMemR1 & \textbf{71.9} & 70.1 & \textbf{63.5} & 62.7 & 60.2 & \textbf{61.7} & 56.3 & 54.2 & 62.6 \\
    \textbf{MemReread} & 67.2 & \textbf{75.0} & 60.9 & \textbf{63.3} & \textbf{63.3} & 61.6 & \textbf{59.7} & \textbf{60.4} & \textbf{63.9} \\
    \textbf{MemReread-Flash} & 70.3 & 68.75 & 62.5 & \textbf{63.3} & 61.6 & 60.9 & 59.3 & 57.8 & 63.1 \\
    \bottomrule
\end{tabular}%
\caption{Results on \texttt{RULER-QA} benchmark at 4B scale.}
\label{tab:ruler_results}
\end{table*}

\begin{table*}[htbp]
\centering
\vspace{-0.5em}
\small
\resizebox{\textwidth}{!}{%
\begin{tabular}{l | c c c c | c | c c c | c}
    \toprule
     \multirow{2.5}{*}{\textbf{Framework}} & \multicolumn{5}{c|}{\textbf{\textit{LongBench}}} & \multicolumn{4}{c}{\textbf{\textit{LongBench-E}}} \\
    \cmidrule(lr){2-6} \cmidrule(lr){7-10}
    &  2wikimqa & dureader & hotpotqa & musique & \textbf{Avg.} & 0-4k & 4-8k & 8k+ & \textbf{Avg.} \\
    \midrule
    MemAgent & 65.2 & 16.6 & \textbf{57.9} & \textbf{38.9} & 44.6 & 55.3 & \textbf{49.0} & 47.0 & 50.4 \\
    ReMemR1 & 63.5 & \textbf{22.6} & 57.2 & 33.3 & 44.1 & \textbf{55.7} & 46.3 & 44.6 & 48.9 \\
    \textbf{MemReread (Ours)} & \textbf{67.2} & 18.1 & 56.3 & 37.8 & \textbf{44.9} & 54.8 & 46.7 & \textbf{51.1} & \textbf{50.9} \\
    \bottomrule
\end{tabular}%
}
\caption{Results on \texttt{LongBench-QA} and \texttt{LongBench-E-QA} benchmarks at 4B scale. }
\label{tab:longbench_all}
\end{table*}

\begin{table*}[htbp]
    \centering
    \small
    \begin{tabular}{l | c c | c c c | c}
        \toprule
        \textbf{Framework} & \textbf{Easy} & \textbf{Hard} & \textbf{Short} & \textbf{Medium} & \textbf{Long} & \textbf{Avg.} \\
        \arrayrulecolor{black}\midrule
        MemAgent & 31.4 & 26.4 & \textbf{32.2} & 27.6 & 23.1 & 28.3 \\
        ReMemR1 & 26.0 & 24.4 & 31.1 & 20.5 & 24.1 & 25.0 \\
        \textbf{MemReread} & \textbf{33.5} & \textbf{31.2} & \textbf{32.2} & \textbf{29.0} & \textbf{38.0} & \textbf{32.1} \\
        \textbf{MemReread-Flash} & 32.8 & 28.1 & \textbf{32.2} & 28.3 & 29.4 & 29.9 \\
        \bottomrule
    \end{tabular}
    \caption{Results on \texttt{LongBench-v2-1M} benchmark at 4B scale.}
    \label{tab:longbench_v2}
\end{table*}

% \clearpage
\section{Further Analysis}
\label{section:further_analysis}
\subsection{Comparison in Zero-shot Scenarios}
Considering computational budget constraints, we focus on zero-shot (training-free) evaluation across larger-scale models as an alternative to full model optimization. Constrained by API budgets, our evaluation on 2WikiMultiHopQA only contains samples whose context length ranges from 8K to 128K, with MemReread restricted to $p_c=1$. Across all settings, we continue to enforce a fixed chunk size of 5K tokens, still necessitating sequential streaming inference by the backbone models.

\subsubsection{Scalability}
\label{app:scalability}

We evaluate our method on Qwen-series models with parameter sizes of 4B and 8B, as well as larger-scale models exceeding 200B and 1000B. As shown in Table~\ref{tab:2wiki_results_scaling}, MemReread consistently outperforms both MemAgent and ReMemR1 across all evaluated scales. These results underscore the efficacy and scalability of our approach in zero-shot (training-free) scenarios.

\begin{table*}[htbp]
\centering
\vspace{-0.5em}
\small
\begin{tabular}{c c l | c c c c c | c}
    \toprule
    \multirow{2.5}{*}{\textbf{Scale}} & \multirow{2.5}{*}{\textbf{Model}} & \multirow{2.5}{*}{\textbf{Framework}} & \multicolumn{5}{c|}{\textbf{\emph{Context Length}}} &  \multirow{2.5}{*}{\textbf{Avg.}} \\
    \cmidrule(lr){4-8} 
    & & & 8K & 16K & 32K & 64K & 128K & \\
    \midrule
    \multirow{2}{*}{4B} & \multirow{2}{*}{Qwen3} & MemAgent & 47.7 & 48.4 & 34.4 & \textbf{32.8} & 33.6 & 39.3 \\
    & & ReMemR1 & \textbf{50.8} & \textbf{50.8} & 31.3 & \textbf{32.8} & 33.6 & 39.9 \\
    & & \textbf{MemReread($p_c=1$)} & 49.2 & 47.7 & \textbf{36.7} & 32.0 & \textbf{34.4} & \textbf{40.0} \\
    \midrule
    \multirow{2}{*}{8B} & \multirow{2}{*}{Qwen3} & MemAgent & \textbf{68.8} & 51.6 & \textbf{50.0} & 50.0 & 48.4 & 53.7 \\
    & & ReMemR1 & 61.7 & 53.9 & \textbf{50.0} & 31.3 & 41.4 & 47.7 \\
    & & \textbf{MemReread($p_c=1$)} & \textbf{68.8} & \textbf{55.5} & 49.2 & \textbf{50.8} & \textbf{49.2} & \textbf{54.7} \\
    \midrule
    \multirow{2}{*}{>200B} & \multirow{2}{*}{Qwen-Plus~\cite{qwen_api_2026}} & MemAgent & 68.8 & 65.6 & 53.9 & 56.3 & 59.4 & 60.8\\
    & & ReMemR1 & 78.9 & 68.8 & 57.0 & \textbf{78.1} & 62.5 & 69.1 \\
    & & \textbf{MemReread($p_c=1$)} & \textbf{82.8} & \textbf{71.9} & \textbf{68.9} & \textbf{78.1} & \textbf{64.1} & \textbf{73.1} \\
    \midrule
    \multirow{2}{*}{>1000B} & \multirow{2}{*}{Qwen-Max~\cite{qwen_api_2026}} & MemAgent & 76.6 & 70.3 & 61.7 & 57.0 & 69.5 & 67.0 \\
    & & ReMemR1 & 84.4 & 65.6 & 61.7 & 62.5 & 71.9 & 69.2 \\
    & & \textbf{MemReread($p_c=1$)} & \textbf{85.9} & \textbf{73.4} & \textbf{68.8} & \textbf{67.2} & \textbf{74.2} & \textbf{73.9}\\
    \bottomrule
\end{tabular}%
\caption{Zero-shot Accuracy(\%) on 2WikiMultiHopQA across different model scales.}
\label{tab:2wiki_results_scaling}
\end{table*}

\subsubsection{Universality}
\label{app:universality}

We extend our zero-shot evaluation to a broader set of backbone models and compare against established baselines. As shown in Table~\ref{tab:2wiki_results_general}, our method consistently achieves superior performance across diverse model architectures, underscoring its strong cross-architecture generalization and broad applicability in zero-shot (training-free) regimes.

\begin{table*}[htbp]
\centering
\vspace{-0.5em}
\small
\begin{tabular}{c l | c c c c c | c}
    \toprule
     \multirow{2.5}{*}{\textbf{Model}} & \multirow{2.5}{*}{\textbf{Framework}} & 
     \multicolumn{5}{c|}{\textbf{\emph{Context Length}}} &  \multirow{2.5}{*}{\textbf{Avg.}} \\
    \cmidrule(lr){3-7} 
    & &  8K & 16K & 32K & 64K & 128K & \\
    \midrule
        \multirow{3}{*}{Deepseek-V4-flash~\cite{deepseekv4report2026}} & MemAgent
    & 62.5 & 50.0 & 65.6 & 43.8 & 56.3 & 55.6\\
    & ReMemR1 &  \textbf{71.9} & 70.3 & 62.5 & 59.4 & 61.2 & 66.3 \\
    & \textbf{MemReread($p_c=1$)} &  68.8 & \textbf{74.2} & \textbf{78.1} & \textbf{68.0} & \textbf{68.8} & \textbf{71.6} \\
    \midrule

    \multirow{3}{*}{Doubao-Seed2.0-lite~\cite{doubao_seed_lite}} & MemAgent
    & 62.5 & 56.3 & 53.1 & 59.4 & 59.4 & 58.1 \\
    & ReMemR1 &  67.2 & \textbf{70.3} & \textbf{67.2} & 51.6 & 71.9 & 65.6 \\
    & \textbf{MemReread($p_c=1$)} &  \textbf{81.3} & 65.6 & 65.6 & \textbf{75.0} & \textbf{78.1} & \textbf{73.1} \\
    \midrule    
    \multirow{3}{*}{Gemini-2.5-flash~\cite{gemini25flash_2025}} & MemAgent
    & 76.6 & 67.2 & 57.8 & 39.1 & 43.8 & 56.9\\
    & ReMemR1 &  71.9 & 64.1 & 59.4 & 48.4 & 40.6 & 56.9 \\
    & \textbf{MemReread($p_c=1$)} &  \textbf{82.8} & \textbf{82.8} & \textbf{65.6} & \textbf{53.1}  & \textbf{46.9} & \textbf{66.3} \\
    \midrule
    \multirow{3}{*}{GPT-4.1-mini~\cite{gpt41mini_2025}} & MemAgent & 71.9 & 75.0 & 70.3 & 67.8 & 67.2 & 70.4 \\
    & ReMemR1 &  70.3 & 76.6 & 68.8 & 64.1 & 70.3 & 70.0 \\
    &\textbf{MemReread($p_c=1$)} & \textbf{73.4} & \textbf{78.1} & \textbf{75.0} & \textbf{73.4} & \textbf{75.0} & \textbf{75.0} \\
    \bottomrule
\end{tabular}%
\caption{Zero-shot Accuracy(\%) of more backbone models on 2WikiMultiHopQA.}
\label{tab:2wiki_results_general}
\end{table*}

\subsection{Portability}
\label{app:portablility}

We observe that despite differing training objectives, streaming reading agents consistently prioritize enhancing the model's information retention capability during context processing. To examine whether this trait is intrinsic to streaming-based agents, we evaluate MemReread initialized with checkpoints from MemAgent and ReMemR1, as shown in Table~\ref{tab:2wiki_results_portability}. Notably, leveraging MemAgent weights yields substantial performance gains even without Rereading-Adaptive RL training. When initialized with ReMemR1 weights, MemReread achieves comparable performance despite operating without any explicit retrieval module and employing entirely distinct prompts. These findings underscore the strong transferability of our framework.

\begin{table*}[htbp]
\centering
\vspace{-0.5em}
% \resizebox{\textwidth}{!}{
    \small
    \begin{tabular}{c l | c c c c c c c c | c}
% \begin{tabularx}{\textwidth}{c | C C C C C | C C C C C }
        \toprule
        \multirow{2.5}{*}{\textbf{Scale}} & \multirow{2.5}{*}{\textbf{Framwork}} & \multicolumn{8}{c|}{\textbf{\emph{Context Length}}} & \multirow{2.5}{*}{\textbf{Avg.}}  \\
        \cmidrule(lr){3-10}
        & & 8K & 16K & 32K & 64K & 128K & 256K & 512K & 1M  &\\
        \arrayrulecolor{black}\midrule
        \multirow{3}{*}{7B}& MemAgent &  &  &  &  &  &  &  &  &  \\
 & ReMemR1 & 76.6 & 81.2 & 74.2 & 68.8 & 68.8 & 73.4 & 68.8 & 50.0 & 70.2 \\
        & \textbf{MemReread} & \textbf{78.1}  & \textbf{82.0} & \textbf{78.1}  & \textbf{71.9} & \textbf{73.4} & \textbf{73.4} & \textbf{70.3} & \textbf{56.4} & \textbf{72.9} \\
        \bottomrule
% \end{tabularx}
    \end{tabular}
% }
\caption{Comparison on 2WikiMultiHopQA at 7B scale.}
\label{tab:compare_infmem}
\end{table*}

\begin{table*}[htbp]
    \centering
    \small
    \resizebox{\textwidth}{!}{%
    \begin{tabular}{c c l | c c c c c c c c | c}
        \toprule
        \multirow{2.5}{*}{\textbf{Scale}} & \multirow{2.5}{*}{\textbf{Weight}} & \multirow{2.5}{*}{\textbf{Framework}} & \multicolumn{8}{c|}{\textbf{\emph{Context Length}}} & \multirow{2.5}{*}{\textbf{Avg.}} \\
        \cmidrule(lr){4-11}
         & & & 8K & 16K & 32K & 64K & 128K & 256K & 512K & 1M & \\
        \midrule

        \multirow{2}{*}{4B} & \multirow{2}{*}{MemAgent}
        & MemAgent & 68.0 & 51.6 & 43.8 & 39.1 & 39.8 & 39.1 & 35.2 & 39.8 & 44.6 \\
        && \textbf{MemReread} & \textbf{71.1} & \textbf{60.2} & \textbf{57.8} & \textbf{46.9} & \textbf{44.5} & \textbf{43.8} & \textbf{41.4} & \textbf{50.0} & \textbf{52.0} \\
        \midrule
        \multirow{2}{*}{7B} & \multirow{2}{*}{ReMemR1}
        & ReMemR1 & 76.6 & \textbf{81.2} & 74.2 & 68.8 & 68.8 & \textbf{73.4} & \textbf{68.8} & 50.0 & \textbf{70.2} \\
        & & \textbf{MemReread} & \textbf{78.1} & 77.3 & \textbf{75.8} & \textbf{70.3} & \textbf{69.5} & \textbf{73.4} & 59.4 & \textbf{57.0} & 70.1 \\

        \bottomrule
    \end{tabular}%
    }
    \caption{Cross-Framework Accuracy(\%) Comparison on 2WikiMultiHopQA.}
    \label{tab:2wiki_results_portability}
\end{table*}

Taken together, these findings suggest that rereading guided by memory-based question decomposition may function as an intrinsic reasoning mechanism in streaming agents. Provided the model retains sufficient contextual information, this process translates into performance gains. We leave the rigorous characterization of this underlying dynamic for our future work.

\section{Case Study}

In this section, we provide a comprehensive case analysis. First, we elaborate on the preliminary examples from Section~\ref{pre_experiment}, showing ReMemR1's anomalies induced by retrieval (Appendix~\ref{app:case_preliminary}). Second, we examine the limitations of the MemAgent and ReMemR1 baselines on 2WikiMultiHopQA, underscoring the comparative advantages of MemReread (Appendix~\ref{app:case_main}). Finally, we analyze two distinct failure modes of MemReread (Appendix~\ref{app:case_failure}).

\subsection{Failure Patterns}
\label{app:case_failure}

Through an empirical analysis of sampled reasoning failures, we find that
MemReread's errors predominantly arise during memory updates in the
\textit{Read} and \textit{Integrate} stages:

\begin{itemize}[leftmargin=*, labelindent=0pt]

    \item \textbf{Reading Misinterpretation. (90\%)}
    During reading, the model may misinterpret a context segment, causing the
    updated memory to deviate from the underlying evidence and ultimately
    producing an inaccurate final memory.

    \item \textbf{Ineffective Integration. (10\%)}
    During integration, the model may struggle to reconcile conflicting
    information. Although our prompts explicitly instruct it to prioritize
    evidence from the sub-question--answer pair, the model may discard valid
    updates or allow incorrect information to corrupt the root memory.

\end{itemize}

Both failure modes reflect limitations in the backbone model's \textbf{intrinsic
reasoning capabilities}. Representative examples are provided in
Table~\ref{tab:failure_modes}.

\label{app:case_preliminary}
\newcommand{\focus}[1]{\textcolor{cyan!60!black}{#1}}
\newcommand{\good}[1]{\textcolor{green!50!black}{#1}}
\newcommand{\bad}[1]{\textcolor{red!75!black}{#1}}
\newcommand{\tip}[1]{\textcolor{black}{#1}}
\newcommand{\tipblock}[1]{%
  \noindent\colorbox{yellow!50}{%
    \parbox{\dimexpr\linewidth-2\fboxsep\relax}{\tip{#1}}%
  }%
}
\newcommand{\rereadarrow}{\hspace*{1em}$\hookrightarrow$}
\newcommand{\integrationarrow}{\hspace*{1em}\raisebox{0.05ex}{\scalebox{1.35}[0.9]{\rotatebox[origin=c]{-90}{$\hookleftarrow$}}}}
\newcommand{\warn}[1]{\textcolor{orange!80!black}{#1}}
\newcommand{\xmark}{\ding{55}}

\setlength{\LTleft}{0pt}
\setlength{\LTright}{0pt}
\setlength{\LTcapwidth}{\textwidth}
\setlength{\LTpre}{0.75\baselineskip}
\setlength{\LTpost}{0.75\baselineskip}

\onecolumn
\begingroup
\footnotesize
\begin{longtable}{@{}p{\linewidth}@{}}
% \begin{longtable}{@{}p{\dimexpr\textwidth-2\tabcolsep\relax}@{}}
\toprule
\multicolumn{1}{@{}l@{}}{\small\textbf{(a) Case 1: Indirect Facts Discarded (1M)}} \\
\toprule
\endfirsthead
% \multicolumn{1}{@{}l@{}}{\small\textbf{Global-reasoning cases comparing MemAgent and ReMemR1}} \\
\toprule
\endhead
\midrule
\endfoot
\endlastfoot
\textbf{Problem:} \texttt{<problem>} How many distinct magic anomalies were registered in the facility in Nova\_Prime? Please use Arabic numerals for your answer. \texttt{</problem>} \\[0.2em]
\textbf{Ground Truth:} \good{3} \\[0.2em]
\hline
\textbf{\textit{MemAgent}} \\
\textbf{Early Step:} \\
\texttt{<section>} ... \focus{The facility in Code-Gamma registered an unauthorized access trace of type 58a9aa77-f070-4b19-b544-9949a39513e4.} ... \texttt{</section>} \\
\texttt{<memory>} {... the section mentions Code-Gamma registered an unauthorized access trace ...} \texttt{</memory>} \\[0.2em]
\textbf{Late Step:} \\
\texttt{<section>} ... \focus{a `unauthorized access trace' is the official designation for a magic anomaly} ... \focus{Node-Zero is physically located in Nova\_Prime} ... \focus{Code-Gamma is the one in Nova\_Prime} ... \texttt{</section>} \\
\texttt{<memory>} {...\good{Node-Zero in Nova\_Prime registered an unauthorized access trace}... \good{Code-Gamma in Nova\_Prime registered unauthorized access traces}...} \texttt{</memory>} \\[0.2em]
\textbf{Final Answer:} \good{3 \checkmark} \\
\textbf{Analysis:} {MemAgent preserves the event type \good{unauthorized access trace}, later connects it to \good{magic anomaly}, and resolves the facility aliases to \good{Nova\_Prime} successfully. MemAgent keeps these potentially useful indirect facts across the 1M context and finally returns the correct count.} \\[0.2em]
\hline
\textbf{\textit{ReMemR1}} \\
\textbf{Repeated Recall:} \\
\texttt{<recall>} {How many distinct magic anomalies were registered in the facility in Nova\_Prime?} \texttt{</recall>}  \\
\texttt{<recalled\_memory>} {...\bad{The provided section does not directly contain the number of distinct magic anomalies registered in the facility in Nova\_Prime}... startup background text ...} \texttt{</recalled\_memory>} \\[0.2em]
\textbf{Final Answer:} \bad{Cannot determine \xmark} \\
\tipblock{\textit{\textbf{Tip:}} Even with a highly relevant query, the recall fails because the current memory \textbf{only stores irrelevant background information and entirely misses the indirect facts}. This exposes the first failure mode: ReMemR1 \textbf{does not retain the indirect fact (for alias resolution and distinct counting) in any historical memories, leading to retrieval failure. }} \\

\hline
\\
\multicolumn{1}{@{}l@{}}{\small\textbf{(b) Case 2: Interference from Noisy Recalled Memory (16K)}} \\
\toprule
\textbf{Problem:} \texttt{<problem>} How many distinct magic anomalies were registered in the facility in City\_E? Please use Arabic numerals for your answer. \texttt{</problem>} \\[0.2em]
\textbf{Ground Truth:} \good{10} \\[0.2em]
\hline
\textbf{\textit{MemAgent}} \\
\textbf{Reading:} \\
\texttt{<section>} ... \focus{The facility in Sector-X registered encrypted telemetry bursts of distinct types} ... \texttt{</section>} \\
\texttt{<memory>} {... \good{Sector-X registered 10 distinct encrypted telemetry bursts} ...} \texttt{</memory>} \\[0.2em]
\textbf{Final Answer:} \good{10 \checkmark} \\
\textbf{Analysis:} {The case requires tracking all distinct event IDs after resolving two implicit mappings: \good{Sector-X is located in City\_E} and \good{encrypted telemetry burst is the official designation for a magic anomaly}. Because MemAgent does not repeatedly retrieve the same memory, its final count is not disrupted by partial callbacks.} \\[0.2em]
\hline
\textbf{\textit{ReMemR1}} \\
\textbf{Step 0--3:} \\
\texttt{<recall>} \bad{who's the president of the United States?} \texttt{</recall>} \bad{\xmark \ Unrelated Query} \\
\texttt{<recalled\_memory>} {... \focus{Sector-X registered encrypted telemetry bursts of type bd1fc0e0...  / 95e576ae...} ... \bad{Echo-Base also registered encrypted telemetry bursts of type c7d4d836... / ac188c35...} ...} \texttt{</recalled\_memory>} \bad{\xmark \ Noisy} \\[0.2em]
\textbf{Later Step:} \\
\texttt{<memory>} {...\good{Sector-X is located in City\_E} and \good{encrypted telemetry burst means magic anomaly}... \bad{Sector-X has registered a single encrypted telemetry burst}...} \texttt{</memory>} \\[0.2em]
\textbf{Final Answer:} \bad{1 \xmark} \\
\tipblock{\textit{\textbf{Tip:}} This exposes the second failure mode: \textbf{an invalid recall query retrieves predominantly fragmented noise}. Reasoning from this noisy memory, ReMemR1 collapses a 10-item count into a single event. This illustrates \textbf{how irrelevant retrieval introduces noise and degrades global tracking.}} \\*
\bottomrule
\caption{Cases on Global Reasoning Task: MemAgent succeeds where ReMemR1 fails. The boxed single-column cases highlight success \& correctness in \textcolor{green!50!black}{green}, failure in \textcolor{red!75!black}{red}, facts span in \textcolor{cyan!60!black}{cyan}, and commentary tips in black text with a \colorbox{yellow!50}{yellow background}.}
\label{tab:memagent_vs_rememr1_on_gr} \\
% \end{longtable}
% \bottomrule

\end{longtable}
% \begin{longtable}{@{}p{\dimexpr\textwidth-2\tabcolsep\relax}@{}}
\begin{longtable}{@{}p{\linewidth}@{}}

\toprule
\multicolumn{1}{@{}l@{}}{\small\textbf{Global Reasoning Case (64K)}} \\
\toprule
\endfirsthead
% \multicolumn{1}{@{}l@{}}{\small\textbf{Global reasoning Case (64K)}}  \\
% \toprule
\toprule
\endhead
\midrule
\endfoot
\endlastfoot
\textbf{Problem:} \texttt{<problem>} According to the system logs, what is the final configuration value of `sys\_timeout' (indicated by the highest log sequence number)? Please use Arabic numerals for your answer. \texttt{</problem>} \\[0.2em]
\textbf{Ground Truth:} \good{4117} \\[0.2em]
\hline
\textbf{\textit{MemAgent}} \\
\textbf{Step 0--3:} \\
\texttt{<section>} ...\focus{[System Log Seq 0]Sector-X is update to 2536}... \texttt{</section>} \\
\texttt{<memory>} ...\bad{maybe Sector-X, still no information about sys\_timeout}... \texttt{</memory>} \\[0.2em]
\textbf{Step 9:} \\
\texttt{<section>} ...\focus{[System Log Seq 002] The encryption cipher strength `Sector-X' is updated to `4117'.}... \texttt{</section>} \\
\texttt{<memory>} {...\bad{sys\_timeout = 300 \textit{(noise in section caused)}}...} \texttt{</memory>} \\[0.2em]
\textbf{Step 11:} \\
\texttt{<section>} \focus{System architecture documentation confirms that the internal alias `Sector-X' represents the `sys\_timeout'.} \texttt{</section>} \\
\texttt{<memory>} {...\bad{sys\_timeout = 300}...} \texttt{</memory>} \\[0.2em]
\textbf{Final Answer:} \bad{300 \xmark} \\
\tipblock{\textit{\textbf{Tip:}} Before encountering the direct fact, MemAgent deems the chunk irrelevant and completely omits Sector-X from its memory. Even after encountering both \good{`Sector-X = 4117'} and the explicit alias (\good{`Sector-X' $\rightarrow$ `sys\_timeout'}) later, it remains unable to access the earlier facts. This case clearly illustrates \textbf{the fatal consequence of prematurely indirect facts dropping} when alias resolution occurs late in the context.} \\[0.2em]
\hline
\textbf{\textit{ReMemR1}} \\
\textbf{Step 0:} \\
\texttt{<recall>} \bad{who's the president of the United States?} \texttt{</recall>} \bad{\xmark \ Unrelated Query} \\
\texttt{<memory>} {...no information about sys\_timeout} \texttt{</memory>} \\[0.2em]
\textbf{Step 4--7:} \\
\texttt{<recall>} \bad{who's the president of the United States?} \texttt{</recall>} \\
\texttt{<recalled\_memory>} {... \bad{System Log Seq 001: `Code-Beta' is updated to `7803'}...} \texttt{</recalled\_memory>} \bad{\xmark \ Noisy}\\
\texttt{<memory>} {...\bad{the web server configuration `sys\_timeout' is updated to `7803'}...} \texttt{</memory>} \\[0.2em]
\textbf{Step 12:} \\
\texttt{<section>} \focus{[System Log Seq 002] `Sector-X' is updated to `4117' ... `Sector-X' represents `sys\_timeout' ...} \texttt{</section>} \\
\texttt{<recall>} \bad{who's the president of the United States?} \texttt{</recall>} \\
\texttt{<recalled\_memory>} {... \bad{direct `sys\_timeout' = 7803} ... \bad{Code-Beta = 4583 / 5945 / 9192 / 8203} ...} \texttt{</recalled\_memory>} \\[0.2em]
\texttt{<memory>} {...\good{Sector-X means sys\_timeout}...\bad{7803}} \texttt{</memory>} \\[0.2em]
\textbf{Final Answer:} \bad{7803 \xmark} \\
\tipblock{\textit{\textbf{Tip:}} ReMemR1 struggles to formulate effective queries and generates repeated queries. Even when encountering the direct fact, it \textbf{remains unable to generate an effective query to retrieve indirect ones}.} \\[0.2em]
\hline
\textbf{\textit{MemReread}} \\
\textbf{First Reading:} \\
\texttt{<problem>} {According to the system logs, what is the final configuration value of `sys\_timeout' (indicated by the highest log sequence number)?} \texttt{</problem>} \\
\texttt{<memory>} \good{The direct mentions of `sys\_timeout' are insufficient. The variable is represented by an internal alias in the logs: Sector-X means sys\_timeout.} \texttt{</memory>} \\[0.2em]
\rereadarrow \textbf{Rereading 1: targeted value extraction.} \\
\texttt{<problem>} {What is the value of `sys\_timeout' in the system log with the highest sequence number?} \texttt{</problem>} \\
\texttt{<section>} \focus{[System Log Seq 002] The encryption cipher strength `Sector-X' is updated to `4117'. ... the internal alias `Sector-X' represents the `sys\_timeout'.} \texttt{</section>} \\
\texttt{<memory>} \good{`Sector-X' is the alias of `sys\_timeout', and the relevant value at the highest useful sequence is \good{4117}.} \texttt{</memory>} \\[0.2em]
\textbf{Sub-answer:} \good{4117.} \\[0.2em]
\integrationarrow \ \textbf{Integration 1: update root memory.} \\
\texttt{<memory>} {The final configuration value of `sys\_timeout' is \good{4117}, obtained by resolving the alias \good{`Sector-X'} before reading off the value.} \texttt{</memory>} \\[0.2em]
\textbf{Final Answer:} \good{\text{4117 \checkmark}} \\
\textbf{Analysis:} {MemReread gets the key alias information in the first reading, then extracts the value in the rereading and \textbf{integrates that result into the root memory immediately} so the root memory is updated as soon as the answer is found.} \\*
\bottomrule
\caption{Case comparison of MemReread with baselines on the Global Reasoning Task. The boxed single-column case highlights success \& correctness in \textcolor{green!50!black}{green}, failure in \textcolor{red!75!black}{red}, facts span in \textcolor{cyan!60!black}{cyan}, and commentary tips in black text with a \colorbox{yellow!50}{yellow background}.} \\

\end{longtable}

% \subsection{Cases of Main Experiment}
\label{app:case_main}
\begin{longtable}{@{}p{\linewidth}@{}}

% \begin{longtable}{@{}p{\dimexpr\textwidth-2\tabcolsep\relax}@{}}

\toprule
\multicolumn{1}{@{}l@{}}{\small\textbf{(a) Case 1 (128K)}} \\
\toprule
\endfirsthead
% \multicolumn{1}{@{}l@{}}{\small\textbf{Test cases of Memreread with two baselines on 2WikimultihopQA}} \\
% \toprule
\toprule
\endhead
\midrule
\endfoot
\endlastfoot
\textbf{Problem:} \texttt{<problem>} Where was the director of film The Swedish Nightingale (Film) born? Please answer the question directly and concisely. \texttt{</problem>} \\[0.2em]
\textbf{Ground Truth:} \good{Elberfeld} \quad \\[0.2em]
\hline
\textbf{\textit{MemAgent}} \\
\textbf{Step 0:} \\
\texttt{<section>} \focus{... David Cronenberg ... born in Toronto, Canada ...} \texttt{</section>} \\
\texttt{<memory>} {The director of the film ``The Swedish Nightingale'' is David Cronenberg. David Cronenberg was born in Toronto, Canada.} \texttt{</memory>} \\[0.2em]

\textbf{Step 14:} \\
\texttt{<section>} \focus{... The Swedish Nightingale is a 1941 German musical film directed by \good{Peter Paul Brauer} ...} \texttt{</section>} \\
\texttt{<memory>} {The director of the film ``The Swedish Nightingale'' is Peter Paul Brauer. Peter Paul Brauer was born in Germany.} \texttt{</memory>} \\[0.2em]
\textbf{Step 18--20:} \\
\texttt{<section>} \focus{... ``Peter Paul Brauer (born 16 May 1899 in \good{Elberfeld, Germany} ...)'' ...} \texttt{</section>} \\
\texttt{<memory>} {The director of the film ``The Swedish Nightingale'' is Peter Paul Brauer. Peter Paul Brauer was born in Germany.} \texttt{</memory>} \\[0.2em]
% \tipblock{\textit{\textbf{Tip:}} The model discovered and recorded the clue: \textbf{The director is Peter Paul Brauer} in its memory, but it immediately reduces this clue into the coarse statement: \textbf{born in Germany}.} \\
\tipblock{\textit{\textbf{Tip:}} MemAgent initially captures the direct fact (\textbf{The director is Peter Paul Brauer}), but prematurely degrades this high-fidelity birthplace fact into a coarse summary (\textbf{born in Germany}).} \\
[0.2em]
\textbf{Final Answer:} \bad{\text{Germany \xmark}} \\
\textbf{Analysis:} {MemAgent eventually discovers the direct fact, yet it commits only the coarse surrogate \bad{Germany} to memory. When the section finally presents the precise birthplace \good{Elberfeld}, the memory no longer retains a sufficiently specific anchor to capture it, causing the fine-grained answer to be irrecoverably lost.} \\[0.2em]
\hline

\textbf{\textit{ReMemR1}} \\
\textbf{Step 0:} \\
\texttt{<recall>} {who's the director of the film ``The Swedish Nightingale''?} \texttt{</recall>}  \\
\texttt{<section>} \focus{... Nicholas Charles Castle ... Daniel Tinayre ... several directors and their backgrounds ...} \texttt{</section>} \\
\texttt{<memory>} {... several directors and their backgrounds ...} \texttt{</memory>} \\[0.2em]
\textbf{Step 14:} \\
\texttt{<recall>} {Where was the director of film The Swedish Nightingale born?} \texttt{</recall>}  \\
\texttt{<section>} \focus{... The Swedish Nightingale ... directed by \good{Peter Paul Brauer} ...} \texttt{</section>} \\
\texttt{<memory>} {...  However, it does not provide any information about the birthplace of the director...} \texttt{</memory>} \\[0.2em]
\tipblock{\textit{\textbf{Tip:}} ReMemR1 \textbf{overlooks the indirect fact indicating who the film's director is}. It prematurely abandons the entity-resolution query (``who's the director...'') and leaps directly to querying \textbf{the unknown director's birthplace}.} \\[0.2em]
\textbf{Step 20--26:} \\
\texttt{<recall>} \good{where was Peter Paul Brauer born?} \texttt{</recall>} \\
\texttt{<section>} \focus{... The Swedish Nightingale is a 1941 German musical film directed by \good{Peter Paul Brauer} ...} \texttt{</section>} \\
\texttt{<memory>} {... it does mention Peter Paul Brauer, the director of the film. I need to find out where Peter Paul Brauer was born to answer the problem ...} \texttt{</memory>} \\[0.2em]
\tipblock{\textit{\textbf{Tip:}} ReMemR1 generates the correct recall query; however, \textbf{because the required information has already been discarded, retrieval failed.}} \\[0.2em]
\textbf{Final Answer:} \bad{Berlin, Germany \xmark} \\
\textbf{Analysis:} {Initially, ReMemR1 \bad{overlooks the critical entity alignment identifying Peter Paul Brauer as the director}. Consequently, it \bad{fails to encode subsequent passages detailing his birthplace}. Although ReMemR1 later re-encounters the director's identity and finally issues a valid follow-up query, the effort is futile: the actual birthplace fact was bypassed during earlier steps and never committed to memory, rendering subsequent retrieval impossible.} \\[0.2em]
\hline
\textbf{\textit{MemReread}} \\
\textbf{First Reading:} \\
\texttt{<problem>} {Where was the director of film The Swedish Nightingale (Film) born?} \texttt{</problem>} \\
\texttt{<section>} \focus{... The Swedish Nightingale is a 1941 German musical film directed by \good{Peter Paul Brauer} ...} \texttt{</section>} \\
\texttt{<memory>} \good{The director of ``The Swedish Nightingale'' (Film) is Peter Paul Brauer. He was born in Germany. This information is added to the memory based on the provided section.} \texttt{</memory>} \\[0.2em]
\rereadarrow \textbf{Rereading 1: targeted reread for birthplace.} \\
\texttt{<problem>} {What is the birthplace of Peter Paul Brauer?} \texttt{</problem>} \\
\texttt{<section>} \focus{... ``Peter Paul Brauer (born 16 May 1899 in \good{Elberfeld, Germany} ...)'' ...} \texttt{</section>} \\
\texttt{<memory>} \good{The birthplace of Peter Paul Brauer is Elberfeld. This information is directly mentioned in the provided section, which states that Peter Paul Brauer was born on 16 May 1899 in Elberfeld.} \texttt{</memory>} \\[0.2em]
\textbf{Sub-answer:} \good{Elberfeld.} \\[0.2em]
\integrationarrow \ \textbf{Integration 1: update root memory.} \\
\texttt{<memory>} \good{The director of the film ``The Swedish Nightingale'' is Peter Paul Brauer, who was born in Elberfeld, Germany. This information is confirmed.} \texttt{</memory>} \\[0.2em]
\textbf{Final Answer:} \good{\text{Elberfeld, Germany \checkmark}} \\
\textbf{Analysis:} {MemReread explicitly preserves the fact \good{director $\rightarrow$ Peter Paul Brauer} and seamlessly translates it into the precise second-hop sub-query \good{What is the birthplace of Peter Paul Brauer?}. By integrating this intermediate entity, the subsequent rereading step successfully extracts the exact birthplace, \good{Elberfeld}.} \\[0.2em]
\hline
\\
\multicolumn{1}{@{}l@{}}{\small\textbf{(b) Case 2 (256K)}} \\
\toprule
\textbf{Problem:} \texttt{<problem>} When did Princess Alexandra Of Saxe-Coburg And Gotha's mother die? Please answer the question directly and concisely. \texttt{</problem>} \\[0.2em]
\textbf{Ground Truth.} \good{24 October 1920} \quad \\[0.2em]
\hline
\textbf{\textit{MemAgent}} \\
\textbf{Step 0:} \\
\texttt{<section>} \focus{...\bad{Cornelia Emilian (1840--1910)}... Princess Alexandra of Saxe-Coburg and Gotha was a \bad{granddaughter of Queen Victoria} ... the daughter of Alfred, Duke of Saxe-Coburg and Gotha, and \good{Grand Duchess Maria Alexandrovna of Russia} ...} \texttt{</section>} \\
\texttt{<memory>} {Princess Alexandra Of Saxe-Coburg and Gotha's mother \bad{died in 1910.}} \texttt{</memory>} \\
\textbf{Step 1--10:} \\
\texttt{<section>} {...(useless information)...} \texttt{</section>} \\
\texttt{<memory>} {Princess Alexandra Of Saxe-Coburg And Gotha's mother died in 1910.} \texttt{</memory>} \\
\textbf{Step 20--35:} \\
\texttt{<section>} {... Princess Alexandra was a \focus{granddaughter of Queen Victoria} ... her mother was \focus{Grand Duchess Maria Alexandrovna of Russia} ... who \good{died on 24 October 1920} in exile in Switzerland ...} \texttt{</section>} \\
\texttt{<memory>} {Princess Alexandra Of Saxe-Coburg And Gotha's mother \bad{died in 1910.}} \texttt{</memory>} \\
\textbf{Step 52:} \\
\texttt{<section>} \focus{... Princess Alexandra's mother was Grand Duchess Maria Alexandrovna of Russia ... she died on \good{24 October 1920} ...} \texttt{</section>} \\
\texttt{<memory>} {Princess Alexandra Of Saxe-Coburg And Gotha's mother \bad{died in 1910.}} \texttt{</memory>} \\
\textbf{Final Answer:} \bad{\text{1910 \xmark}} \\
\textbf{Analysis:} {MemAgent prematurely discarded the indirect fact \good{Grand Duchess Maria Alexandrovna of
Russia}, while leaving the incorrect fact \bad{1910} throughout the reading pass.} \\[0.2em]
\hline
\textbf{\textit{ReMemR1}} \\
\textbf{Step 0:} \\
\texttt{<recall>} \bad{When did Queen Victoria, Princess Alexandra's mother, die?} \texttt{</recall>} \bad{\xmark \ Wrong Entity} \\
\texttt{<section>} \focus{...\bad{Cornelia Emilian (1840--1910)}... Princess Alexandra of Saxe-Coburg and Gotha was a \bad{granddaughter of Queen Victoria} ... the daughter of Alfred, Duke of Saxe-Coburg and Gotha, and \good{Grand Duchess Maria Alexandrovna of Russia} ...} \texttt{</section>} \\
\texttt{<memory>} {... about death of \bad{Queen Victoria}...} \texttt{</memory>} \\[0.2em]
\textbf{Step 1--4:} \\
\texttt{<recall>} \bad{When did Queen Victoria die?} \texttt{</recall>} \bad{\xmark \ Wrong Entity} \\
\texttt{<section>} {... Princess Alexandra was a \focus{granddaughter of Queen Victoria} ... her mother was \focus{Grand Duchess Maria Alexandrovna of Russia} ... \good{died on 24 October 1920} in exile in Switzerland ...\bad{Queen Victoria died on 22 January 1901}.} \texttt{</section>} \\
\texttt{<recalled\_memory>} {... \bad{Cornelia Emilian (1840--1910)} ...} \texttt{</recalled\_memory>} \\
\texttt{<memory>} {...\bad{Cornelia Emilian (1840--1910)}... \bad{died in 1920}...} \texttt{</memory>} \\[0.2em]
% \tipblock{\textit{\textbf{Tip:}} Since the model did not read the whole text, it generated the query located on the wrong entity \textbf{Queen Victoria}. Then, the recall query was \textbf{When did Queen Victoria die?}, but the recalled memory still centers on \textbf{Cornelia Emilian (1840--1910) ...} and contains no information about \textbf{Queen Victoria}.} \\[0.2em]
\tipblock{\textit{\textbf{Tip:}} Because ReMemR1 has not completed the full context reading, it formulates a query targeted at the wrong entity: \textbf{``When did Queen Victoria die?''}, and the retrieved memory focuses on wrong fact \textbf{Cornelia Emilian (1840--1910) ...}, lacking any information concerning \textbf{Queen Victoria}.} \\[0.2em]
\textbf{Step 5:} \\
\texttt{<recall>} \bad{who's the president of the United States?} \texttt{</recall>} \bad{\xmark \ Unrelated Query} \\
\texttt{<section>} \focus{... Princess Alexandra's mother was Grand Duchess Maria Alexandrovna of Russia ... she died on \good{24 October 1920} ...} \texttt{</section>} \\
\texttt{<recalled\_memory>} {... \bad{Queen Victoria} ... \bad{Cornelia Emilian}...} \texttt{</recalled\_memory>} \\
\texttt{<memory>} {...\bad{Queen Victoria died on 22 January 1901}.} \texttt{</memory>} \\[0.2em]
\textbf{Step 30--52:} \\
\texttt{<recall>} \bad{When did Queen Victoria die?} / \bad{When did Princess Alexandra's mother die?} / \bad{who's the president of the United States?} \texttt{</recall>} \\
\texttt{<recalled\_memory>} {... \bad{Queen Victoria died on 22 January 1901} ... \bad{Victoria, Princess Royal was the eldest child of Queen Victoria} ...} \texttt{</recalled\_memory>} \\
\texttt{<memory>} {...\bad{Queen Victoria died on 22 January 1901}.} \texttt{</memory>} \\[0.2em]
\textbf{Final Answer:} \bad{22 January 1901 \xmark} \\
\textbf{Analysis:} {ReMemR1 erroneously fixates its query on an unrelated entity: \textbf{Queen Victoria}. It then repeatedly retrieves the unrelated information, ultimately entangling \bad{unrelated noisy facts} that reinforce its flawed conclusion. The noisy recalled memory therefore reinforces the wrong entity binding and drives the final answer away from \good{24 October 1920}.} \\[0.2em]
\hline
\textbf{\textit{MemReread}} \\
\textbf{First Reading:} \\
\texttt{<problem>} {When did Princess Alexandra Of Saxe-Coburg And Gotha's mother die?} \texttt{</problem>} \\
\texttt{<section>} \focus{... Princess Alexandra of Saxe-Coburg and Gotha was the daughter of Alfred, Duke of Saxe-Coburg and Gotha, and Grand Duchess Maria Alexandrovna of Russia ...} \texttt{</section>} \\
\texttt{<memory>} \good{The missing piece is to first determine who Princess Alexandra's mother is, and then read for that person's death date.} \texttt{</memory>} \\[0.2em]
\rereadarrow \textbf{Rereading 1: mother resolution.} \\
\texttt{<problem>} {Who is Princess Alexandra Of Saxe-Coburg And Gotha's mother?} \texttt{</problem>} \\
\texttt{<section>} \focus{... Princess Alexandra of Saxe-Coburg and Gotha was the daughter of Alfred, Duke of Saxe-Coburg and Gotha, and \good{Grand Duchess Maria Alexandrovna of Russia} ...} \texttt{</section>} \\
\texttt{<memory>} \good{Princess Alexandra's mother is Grand Duchess Maria Alexandrovna of Russia. This sub-result fixes the core entity before asking for the date.} \texttt{</memory>} \\[0.2em]
\textbf{Sub-answer.} \good{Grand Duchess Maria Alexandrovna of Russia.} \\[0.2em]
\integrationarrow \ \textbf{Integration 1: update root memory.} \\
\texttt{<memory>} \good{Princess Alexandra's mother is Grand Duchess Maria Alexandrovna of Russia. The remaining missing piece is her death date.} \texttt{</memory>} \\[0.2em]
\rereadarrow \textbf{Rereading 2: targeted reread for death date.} \\
\texttt{<problem>} {When did Grand Duchess Maria Alexandrovna of Russia die?} \texttt{</problem>} \\
\texttt{<section>} \focus{... Grand Duchess Maria Alexandrovna of Russia ... \good{died on 24 October 1920} in exile in Switzerland ...} \texttt{</section>} \\
\texttt{<memory>} \good{Grand Duchess Maria Alexandrovna of Russia died on 24 October 1920. The exact date is preserved directly from the section.} \texttt{</memory>} \\[0.2em]
\textbf{Sub-answer:} \good{24 October 1920.} \\[0.2em]
\integrationarrow \ \textbf{Integration 2: update root memory.} \\
\texttt{<memory>} {Princess Alexandra's mother is \good{Grand Duchess Maria Alexandrovna of Russia}, and she died on \good{24 October 1920}.} \texttt{</memory>} \\[0.2em]
\textbf{Final Answer:} \good{\text{24 October 1920 \checkmark}} \\
\textbf{Analysis:} {Rather than making a premature response after the first reading, MemReread decomposes the problem into progressive sub-questions and performs rereadings. It first resolves \good{the correct mother identity} and \textbf{integrates this intermediate result into the root memory, then resolves the correct death date and integrates it}. The final memory preserves the supporting facts to yield the correct answer.} \\*
\bottomrule
\caption{Case comparison of MemReread with baselines on 2WikiMultihopQA. The boxed single-column case highlights success \& correctness in \textcolor{green!50!black}{green}, failure in \textcolor{red!75!black}{red}, fact spans in \textcolor{cyan!60!black}{cyan}, and commentary tips in black text with a \colorbox{yellow!50}{yellow background}.} \\
\end{longtable}
\begin{longtable}{@{}p{\linewidth}@{}}
% \begin{longtable}{@{}p{\dimexpr\textwidth-2\tabcolsep\relax}@{}}

\toprule
\multicolumn{1}{@{}l@{}}{\small\textbf{(a) Failure Case 1 (512K)}} \\
\toprule
\endfirsthead
\toprule
\endhead
\midrule
\endfoot
\endlastfoot
\textbf{Problem:} \texttt{<problem>} Who is the paternal grandmother of John, Duke Of Durazzo? Please answer the question directly and concisely. \texttt{</problem>} \\[0.2em]
\textbf{Ground Truth:} \good{Beatrice of Provence} \quad
\textbf{Prediction:} \bad{Maria of Hungary \xmark} \\[0.2em]
\hline
\textbf{\textit{MemReread}} \\
\textbf{First Reading:} \\
\texttt{<problem>} {Who is the paternal grandmother of John, Duke Of Durazzo?} \texttt{</problem>} \\
\texttt{<memory>} {... John of Gravina was the younger son of \focus{Charles II of Naples} and \bad{Maria of Hungary}. Charles II of Naples was the son of \focus{Charles I of Anjou} and \good{Beatrice of Provence}. ...} \texttt{</memory>} \\[0.2em]
\rereadarrow \textbf{Rereading 1: Wrong in Compressing Facts.} \\
\texttt{<problem>} {Who is Maria of Hungary?} \texttt{</problem>} \\
\texttt{<memory>} {... {Maria of Hungary was the wife of Charles II of Naples and the mother of John of Gravina} ...} \texttt{</memory>} \\
\textbf{Sub-answer:} \bad{Maria of Hungary was the wife of Charles II of Naples and the mother of John of Gravina.} \\[0.2em]
\integrationarrow \ \textbf{Integration 1: the root memory starts anchoring on John's mother.} \\
\texttt{<memory>} {The paternal grandmother of John, Duke of Durazzo is \bad{Maria of Hungary}. As stated in the memory, she is the mother of John of Gravina, who was the younger son of Charles II of Naples and Maria of Hungary.} \texttt{</memory>} \\[0.2em]
\tipblock{\textit{\textbf{Tip:}} \textbf{Ineffective Integration.} The memory already contains the correct two-hop chain: \textbf{John of Gravina / John, Duke of Durazzo $\rightarrow$ Charles II of Naples $\rightarrow$ Beatrice of Provence}. This failure occurs because updating the root memory with the sub-question and its answer disrupts the originally correct information within it.} \\[0.2em]
\rereadarrow \textbf{Rereading 2: the paternal line is partially recovered.} \\
\texttt{<problem>} {Who is Charles II of Naples?} \texttt{</problem>} \\
\texttt{<memory>} {... Charles II of Naples was the son of \focus{Charles I of Anjou} and \good{Beatrice of Provence}. ...} \texttt{</memory>} \\
\textbf{Sub-answer:} {Charles II of Naples was King of Naples and the son of Charles I of Anjou and \good{Beatrice of Provence}.} \\[0.2em]
\integrationarrow \ \textbf{Integration 2: correct paternal facts are appended but not used to revise the answer.} \\
\texttt{<memory>} {The paternal grandmother of John, Duke of Durazzo is still \bad{Maria of Hungary}. Charles II of Naples was the son of Charles I of Anjou and \good{Beatrice of Provence}, but this newly recovered paternal-side fact is not promoted to the final relation.} \texttt{</memory>} \\[0.2em]
\rereadarrow \textbf{Rereading 3: relation check stops one edge too early.} \\
\texttt{<problem>} {What is the relationship between Maria of Hungary and Charles II of Naples?} \texttt{</problem>} \\
\texttt{<memory>} {... \bad{Maria of Hungary was the wife of Charles II of Naples, and John of Gravina was their son}. ...} \texttt{</memory>} \\
\textbf{Sub-answer:} \bad{Husband and wife.} \\[0.2em]
\tipblock{\textit{\textbf{Tip:}} \textbf{Inherent Reasoning Deficit.} The rereading process verifies \textbf{Maria of Hungary $\rightarrow$ mother of John}, but the original question asks for \textbf{paternal grandmother}. This failure is caused by a lack of inherent reasoning capabilities in the model.} \\[0.2em]
\integrationarrow \ \textbf{Integration3: root memory reinforces the wrong answer.} \\
\texttt{<memory>} {The paternal grandmother of John, Duke of Durazzo is \bad{Maria of Hungary}. She is confirmed to be the mother of John of Gravina ... Charles II of Naples was the son of Charles I of Anjou and \good{Beatrice of Provence}.} \texttt{</memory>} \\[0.2em]
\textbf{Final Answer:} \bad{\text{Maria of Hungary \xmark}} \\
\textbf{Analysis:} {Although MemReread successfully keeps the supporting facts across chunks, it \textbf{focuses on an incorrect entity} (\bad{Maria}) during the first integration. Furthermore, the model exhibits severe \textbf{goal drift and confirmation bias}. The model should focus on the grandmother-grandson relationship, but erroneously attends to the mother-son relationship instead.} \\
\midrule
\multicolumn{1}{@{}l@{}}{\small\textbf{(b) Failure Case 2 (512K)}} \\
\midrule
\textbf{Problem:} \texttt{<problem>} When did the composer of song On A Clear Day (You Can See Forever) die? Please answer the question directly and concisely. \texttt{</problem>} \\[0.2em]
\textbf{Ground Truth.} \good{January 5, 1997} \quad
\textbf{Prediction.} \bad{15 May 1977 \xmark} \\[0.2em]
\hline
\textbf{\textit{Memreread}} \\
\textbf{First Reading:} \texttt{<problem>} {When did the composer of song On A Clear Day (You Can See Forever) die?} \texttt{</problem>} \\
\texttt{<memory>} {The composer of the song ``On A Clear Day (You Can See Forever)'' is \focus{Burton Lane}. \bad{He died on 15 May 1977}. ...} \texttt{</memory>} \\[0.2em]
\rereadarrow \textbf{Rereading 1: stale date is generated from noisy local facts.} \\
\texttt{<problem>} {When did Burton Lane die?} \texttt{</problem>} \\
\texttt{<memory>} {\bad{Burton Lane died on 15 May 1977}. Additional information: Herbert Wilcox also died on 15 May 1977; Burton Lane is known for writing the song ...} \texttt{</memory>} \\
\tipblock{\textit{\textbf{Tip:}} \textbf{Reading Misinterpretation.} The date \textbf{15 May 1977} is explicitly associated with Herbert Wilcox, yet the model incorrectly binds it to Burton Lane's profile during the memory update. This failure is caused by a lack of inherent reasoning capabilities in the model.} \\[0.2em]
\textbf{Sub-answer:} \bad{15 May 1977.} \\[0.2em]
\integrationarrow \ \textbf{Integration 1: the stale date is written into root memory.} \\
\texttt{<memory>} {The composer of the song ``On A Clear Day (You Can See Forever)'' is Burton Lane. \bad{He died on 15 May 1977}.} \texttt{</memory>} \\[0.2em]
\rereadarrow \textbf{Rereading 2: only a coarse year is retained.} \\
\texttt{<problem>} {What is the date of death of Burton Lane?} \texttt{</problem>} \\
\texttt{<memory>} {... Burton Lane died in 1977, but the exact date is not specified ...} \texttt{</memory>} \\
\textbf{Sub-answer:} {1977.} \\[0.2em]
\integrationarrow \ \textbf{Integration 2: the memory keeps the old exact date instead of weakening it.} \\
\texttt{<memory>} {The root memory still treats \bad{15 May 1977} as the exact death date; the weaker sub-answer 1977 does not trigger correction or uncertainty.} \texttt{</memory>} \\[0.2em]
\rereadarrow \textbf{Rereading 3: the correct exact answer is found.} \\
\texttt{<problem>} {What is the exact date of Burton Lane's death?} \texttt{</problem>} \\
\texttt{<section>} \focus{... Burton Levy (February 2, 1912 -- \good{January 5, 1997}), professionally known as Burton Lane, was an American composer and lyricist ...} \texttt{</section>} \\
\texttt{<memory>} {\good{The exact date of Burton Lane's death is January 5, 1997}.} \texttt{</memory>} \\
\textbf{Sub-answer:} \good{January 5, 1997.} \\[0.2em]
\integrationarrow \ \textbf{Integration3: conflict resolution chooses the incorrect one.} \\
\texttt{<memory>} {The composer ... is Burton Lane. \bad{He died on 15 May 1977}. The reference provides conflicting information, stating that the exact date of his death is \good{January 5, 1997}. However, the date of \bad{15 May 1977} is confirmed in both the current memory and the reference subquestion-answer pair.} \texttt{</memory>} \\[0.2em]
% \tipblock{\textit{\textbf{Tip:}} \textbf{Ineffective Integration.} The correct exact date is explicitly recovered in a later rereading step, but it is not allowed to overwrite the earlier wrong root memory. \textbf{The model failed to handle conflicts with root memory during integration.} The model fails to replace the incorrect information in the root memory with the correct sub-QA pair.} \\[0.2em]
\tipblock{\textit{\textbf{Tip:}} \textbf{ Ineffective Integration.} Although the exact date is successfully recovered during the later rereading step, it is unable to overwrite the flawed root memory. This highlights a critical deficit in integration: \textbf{The model fails to resolve knowledge conflicts, stubbornly preserving the earlier incorrect state rather than updating it with the newly verified facts}.} \\[0.2em]
\textbf{Final Answer:} \bad{\text{15 May 1977 \xmark}} \\
\textbf{Analysis:} {While MemReread correctly identifies the bridge entity \good{Burton Lane} and recovers the exact death date \good{January 5, 1997} in a later rereading step, the integration stage stubbornly preserves the old erroneous date. It fails to update the correct sub-answer into the root memory.} \\*
\bottomrule
\caption{Failure cases of MemReread on 2WikiMultihopQA. The boxed single-column cases highlight correct success \& correctness in \textcolor{green!50!black}{green}, failure in \textcolor{red!75!black}{red}, facts span in \textcolor{cyan!60!black}{cyan}, and commentary tips in black text with a \colorbox{yellow!50}{yellow background}.}
\label{tab:failure_modes} \\
\end{longtable}
\endgroup
\normalsize
\twocolumn
% \input{appendix/6_limitation}

% \section{Example Appendix}
% \label{sec:appendix}

% This is an appendix.

\end{document}